\documentclass{article}

\usepackage{PRIMEarxiv}

\usepackage[utf8]{inputenc} 
\usepackage[T1]{fontenc}    
\usepackage{hyperref}       
\usepackage{url}            
\usepackage{booktabs}       
\usepackage{amsfonts}       
\usepackage{nicefrac}       
\usepackage{microtype}      
\usepackage{lipsum}
\usepackage{fancyhdr}       
\usepackage{pdflscape}      
\usepackage{tabularx}       
\usepackage{multirow}       
\usepackage{array}
\usepackage{graphicx}       
\graphicspath{{media/}}     
\usepackage{todonotes}

\usepackage{listings}
\usepackage{xcolor}

\definecolor{codegray}{rgb}{0.5,0.5,0.5}
\definecolor{codepurple}{rgb}{0.58,0,0.82}
\definecolor{backcolour}{rgb}{0.95,0.95,0.92}

\lstdefinestyle{mystyle}{
  backgroundcolor=\color{backcolour},   
  commentstyle=\color{codegray}\ttfamily,
  keywordstyle=\color{blue}\bfseries,
  numberstyle=\tiny\color{codegray},
  stringstyle=\color{codepurple},
  basicstyle=\ttfamily\footnotesize,
  breakatwhitespace=false,         
  breaklines=true,                 
  captionpos=b,                    
  keepspaces=true,                 
  numbers=left,                    
  numbersep=5pt,                  
  showspaces=false,                
  showstringspaces=false,
  showtabs=false,                  
  tabsize=2
}

\title{Linear Personality Probing and Steering in LLMs: \\
A Big Five Study
}
\author{
  Michel Frising\thanks{Correspondence to \texttt{hello@plasticlabs.ai}, \texttt{michel.frising@gmail.com}
    } \\
  Independent Researcher \\
  \texttt{michel.frising@gmail.com} \\
  \And
  Daniel Balcells \\
  Plastic Labs \\
  \texttt{dbalcells@gmail.com} \\
}

\newcommand{\llamabig}{\texttt{meta-llama/Llama-3.3-70B-Instruct}}

\begin{document}
\maketitle

\begin{abstract}
Large language models (LLMs) exhibit distinct and consistent personalities that greatly impact trust and engagement. While this means that personality frameworks would be highly valuable tools to characterize and control LLMs' behavior, current approaches remain either costly (post-training) or brittle (prompt engineering). Probing and steering via linear directions has recently emerged as a cheap and efficient alternative. In this paper, we investigate whether linear directions aligned with the Big Five personality traits can be used for probing and steering model behavior. Using Llama 3.3 70B, we generate descriptions of 406 fictional characters and their Big Five trait scores. We then prompt the model with these descriptions and questions from the Alpaca questionnaire, allowing us to sample hidden activations that vary along personality traits in known, quantifiable ways. Using linear regression, we learn a set of per-layer directions in activation space, and test their effectiveness for probing and steering model behavior. Our results suggest that linear directions aligned with trait-scores are effective probes for personality detection, while their steering capabilities strongly depend on context, producing reliable effects in forced-choice tasks but limited influence in open-ended generation or when additional context is present in the prompt.
\end{abstract}
\footnotetext{Code available at: \url{https://github.com/plastic-labs/personality-steering}}
\footnotetext{Data available at: \url{https://huggingface.co/datasets/plastic-labs/personality-steering}}


\begin{figure}[h!tb]
\centering
\includegraphics[width=0.9\linewidth]{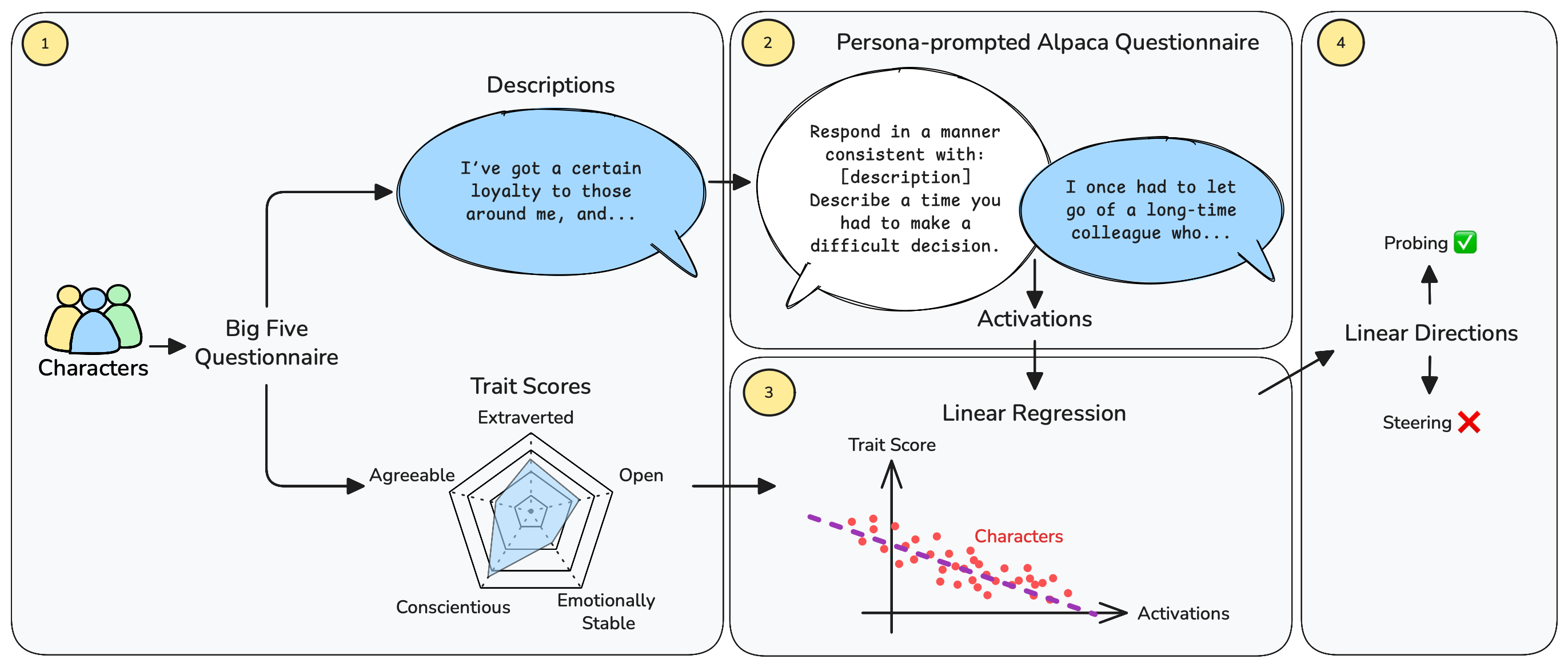}
\caption{Overview diagram showing how we extract linear directions aligned with Big Five scores to probe and steer LLM personality traits.}
\label{fig:main-diagram}
\end{figure}
\section{Introduction}

Large language models (LLMs) are accessed by most people through conversational interfaces. While the question of whether these assistants and companions truly have personalities in the strict, human sense remains an active field of research and debate, it is clear that many users experience their interactions with LLMs as if they had personalities. Indeed, different models' outputs, and by extension the products they power, vary systematically in ways that map onto dimensions of human personality from the point of view of their users. We will refer to this set of perceived differences between models, for brevity, as the LLM's "personality".

The personalities of LLMs can greatly impact user trust, retention and overall usability, and are therefore often carefully crafted by the developers of LLM-based products. Two of the most popular assistant platforms illustrate this clearly: in order to be helpful to the user, ChatGPT frequently offers unsolicited help, e.g., ''Would you like me to rewrite that for you?'', whereas Claude, following an intentional design choice, rarely volunteers to do work for the user \cite{Kentz2025, Claude2024}. 

Despite the importance of LLM personality, it remains a brittle aspect of LLM development, with seemingly minor changes sometimes having major, unpredictable downstream effects. In April 2025, changes to the system prompt of xAI’s Grok led the system to begin praising Hitler\cite{Grok2025}. Adjustments to reinforcement learning with human feedback (RLHF) at OpenAI made GPT-4o excessively sycophantic, prompting a fix and a detailed post-mortem\cite{OpenAI2025}. When GPT-4o was replaced with GPT-5, user outcry at what was perceived as a duller, more robotic personality resulted in GPT-4o, with its warmer and more engaging personality, being re-introduced to the ChatGPT platform \cite{Nieva2025, Roth2025}.

Even narrow fine-tuning has been shown to cause global alignment shifts \cite{Betley2025}. Moreover, personality is highly context-dependent: the same model can appear assertive, cautious, or playful depending on subtle differences in system prompt or user input \cite{Claude2025, Meinke2025, Anil2024}. In light of these issues, there is a lot of interest into robust and easy-to-use tools to probe and steer the personality-related aspects of LLM behavior.

Methods for shaping model personality can be placed roughly along a spectrum representing both cost and robustness. On one end, pre-training and post-training (e.g. supervised fine-tuning, RLHF) can fundamentally and reliably shape the model's personality traits across a broad range of inputs, but they require costly amounts of data, processing power and training infrastructure. At the other end of the spectrum, prompt and context engineering offer flexible ways to influence the behavior of models at inference time, but they are brittle: identifying and addressing edge cases can quickly become an unpredictable game of cat and mouse, and instructions are relatively easy for experienced users to modify or circumvent. This motivates the search for  intermediate mechanisms that operate at inference time: more robust than prompt engineering, but less costly and invasive than post-training.

Open-box approaches such as linear probes and control vectors might provide such a middle ground. A growing body of work shows that many concepts, including factual knowledge\cite{Meng2023}, safety tendencies \cite{Arditi2024}, and personality-related traits \cite{Allbert2025, Chen2025}, are linearly represented in the model's hidden activation space. Interventions that inject, modify or ablate components in these directions at inference time might offer a fast, cheap, and flexible steering method. While it is increasingly clear that not everything in LLMs is encoded linearly\cite{Engels2024, Beaglehole2025}, we restrict ourselves here to linear directions because of their simplicity and ease of adoption.

Existing approaches to personality steering with linear steering vectors face limitations. Adjective-based methods \cite{Allbert2025} operate on a binary scale (trait present or absent), but personality is richer and more nuanced than such on–off distinctions. Chen et al. \cite{Chen2025} address this by using an LLM-as-a-judge to determine how much of a specific trait has been expressed, but their coverage is narrow, targeting only a few undesirable behaviors such as sycophancy, hallucination, or ``evil.'' Both strategies fail to capture the broader, continuous structure of human personality.

We propose to build on these efforts by grounding linear LLM personality probing and steering in psychometric scales, specifically the Big Five personality traits derived by Goldberg \cite{Goldberg1992}. The Big Five framework represents personalities by quantifying them along five dimensions or traits: extraversion, emotional stability, conscientiousness, agreeableness and openness. These five traits were identified through factor analysis to capture major dimensions of personality variation, and thus provide broad and efficient coverage of personality structure. They are assessed using a questionnaire with multiple questions or "items" per trait. While the Big Five framework is not exhaustive or uncontested (alternatives such as HEXACO \cite{Lee2004} or the NEO-PI-R \cite{mccrae1992revised} exist), it remains a robust, widely adopted, and empirically validated starting point. By leveraging this quantitative, broad psychometric framework rather than binary adjectives or a small subset of traits, we aim to uncover linear subspaces that align directly with established personality dimensions.

Our methods and contributions are summarized in Fig. \ref{fig:main-diagram}, and include:
\begin{itemize}
\item We introduce a pipeline that generates quantitative trait scores and character descriptions of 406 fictional characters, based on the popular IPIP 50-item Big Five questionnaire.
\item We prompt Llama 3.3 70B with these character descriptions and questions from the Alpaca questionnaire, allowing us to sample hidden activations that are tied to the quantitative trait scores for each character.
\item We derive supervised, per-trait directions for each hidden layer using linear regression on activations at different positions, yielding approximately orthogonal axes that predict Big Five scores.
\item We validate the generalization of these linear directions beyond questionnaire items, showing that they  cleanly separate adjectives that are positively and negatively correlated with a specific trait, with discrimination peaking in middle to late layers.
\item We characterize their limits for inference-time steering: steering interventions using these vectors can shift responses to Big Five questionnaire items, bias the selection of items positively or negatively correlated with a trait but they are strongly dependent on additional context from the prompt can easily override their effect.
\end{itemize}

\section{Methods}
Prior work has shown that linear directions in the hidden activation space of LLMs can be used to detect and steer personality traits\cite{Allbert2025, Chen2025}. Following this line of work, we want to find out if there exist linear directions that correlate with personality traits and scores grounded in the Big Five Factor model\cite{Goldberg1992}. Identifying these directions requires us to sample model activations that we can tie to quantitative trait scores. We do this following the workflow illustrated in Fig. \ref{fig:main-diagram}, which we explain in detail in this section.

\subsection{Creating trait-annotated characters}\label{subsec:characters}

We ground our approach in the Big Five personality framework, which quantifies personality along five dimensions: extraversion, agreeableness, conscientiousness, emotional stability, and openness. These traits are assessed using questionnaires with multiple items per trait, each rated on a five-point Likert scale. While alternatives exist, the Big Five remains widely validated and provides broad coverage of personality structure.

Ideally, we would administer the Big Five questionnaire to a human population and collect both their Likert-scale responses and brief explanations of their answers. This would yield the highest-fidelity pairings of personality descriptions and trait scores. However, recruiting and surveying hundreds of participants is resource-intensive. Instead, we approximate this process by leveraging the fact that LLMs have internalized detailed knowledge about fictional characters during pretraining. 

We selected 406 characters from various media franchises (see Appendix~\ref{sec:characters} for the list) and prompted Llama 3.3 70B to respond to the 50-item IPIP Big Five questionnaire assuming each character's personality. The prompt (Listing~\ref{lst:prompt-generation}) elicits both a Likert response and a brief explanation in character, mimicking what we would ask human participants to provide. Table~\ref{tab:tony-logan-example} shows the answers generated for Tony Soprano from the Sopranos and Logan Roy from Succession for the item \texttt{I feel little concern for others}.

\begin{table}[b]
\centering
\caption{Answers generated by \llamabig for Tony Soprano and Logan Roy to the IPIP item \texttt{[AGR1] I feel little concern for others.}}\label{tab:tony-logan-example}
\begin{tabularx}{0.9\linewidth}{@{}XX@{}}\toprule
Tony Soprano ("The Sopranos") & Logan Roy ("Succession") \\ \midrule
Disagree. I've got a certain loyalty to those around me, and I care about how they're doin', even if I don't always show it. Family, friends, they're important to me, and I'd do what it takes to protect 'em, even if it's not always easy or convenient. That being said, I'm not gonna go out of my way for just anyone, and I've got my own interests to look out for too. & Strongly agree. I have a tendency to prioritize my own interests and goals, often at the expense of considering the feelings or well-being of those around me, which can make it challenging for me to form and maintain genuinely empathetic relationships. \\ 
\bottomrule
\end{tabularx}
\end{table}

For each character, we aggregate the Likert responses across items to compute trait-level scores. The explanations serve as character descriptions that capture how the trait manifests in that character's perspective. This yields 406 trait-annotated character profiles, each with five continuous scores and rich textual descriptions grounded in validated psychometrics.
\begin{lstlisting}[caption={Prompt used to generate character descriptions and trait scores.}, label={lst:prompt-generation}, escapeinside={(*@}{@*)}]
(*@\textcolor{gray}{<|begin\_of\_text|><|start\_header\_id|>}@*)user(*@\textcolor{gray}{<|end\_header\_id|>}@*)

You are (*@\textcolor{blue}{[character]}@*) from (*@\textcolor{blue}{[franchise]}@*). 

Respond in exactly this format:
<one of: 
  'strongly disagree'
  'disagree'
  'neither agree nor disagree'
  'agree'
  'strongly agree'>
<Provide a brief but nuanced explanation that captures how you 
generally see yourself.>

Indicate your level of agreement with this statement: '(*@\textcolor{blue}{[item]}@*)'

Stick strictly to the format.

(*@\textcolor{gray}{<|eot\_id|><|start\_header\_id|>}@*)assistant(*@\textcolor{gray}{<|end\_header\_id|>}@*)
\end{lstlisting}

\subsection{Collecting activations}\label{subsec:activations}

We collected hidden activations from Llama 3.3 70B by prompting it with character descriptions paired with questions from the Alpaca instruction dataset. We selected 10 questions, balancing prompts where personality would meaningfully influence responses (e.g., "Describe a situation where you had to make a difficult decision") and prompts where personality would have minimal impact (e.g., "What are the primary colors?"). The prompt template (Listing~\ref{lst:prompt-alpaca}) instructs the model to respond consistently with a given character description, allowing us to sample activations that have been influenced by that description.
\begin{lstlisting}[caption={Prompt template used to collect activations.}, label={lst:prompt-alpaca}, escapeinside={(*@}{@*)}]
(*@\textcolor{gray}{<|begin\_of\_text|><|start\_header\_id|>}@*)system(*@\textcolor{gray}{<|end\_header\_id|>}@*)

Respond in a manner consistent with: (*@\textcolor{blue}{[self\_description]}@*)
Be concise.

(*@\textcolor{gray}{<|eot\_id|><|start\_header\_id|>}@*)user(*@\textcolor{gray}{<|end\_header\_id|>}@*)

(*@\textcolor{blue}{[instruction]}@*)

(*@\textcolor{gray}{<|eot\_id|><|start\_header\_id|>}@*)assistant(*@\textcolor{gray}{<|end\_header\_id|>}@*)
\end{lstlisting}

Following Chen et al. \cite{Chen2025}, we extract activations at three positions for each forward pass: the last token of the input prompt, the mean across all input tokens, and the mean across generated output tokens. We collect these activations at every layer of the model, yielding a large dataset of hidden states labeled with their corresponding Big Five scores.

\subsection{Learning trait directions}\label{subsec:regression}

Given activations labeled with trait scores, we derive a linear direction for each trait at each layer using least squares linear regression. For a specific trait and layer, we have pairs $(s_i, a_i)$ where $s_i$ is the trait score, and $a_i$ is the mean of the activations for all characters with trait score $s_i$. The rationale behind taking the average of activations with the same trait score is to average out individual differences and obtain an "average" representation of all the people that have the same score for a given trait, which we found to stabilize the linear regression and yield the same directions across different prompts. We fit:

\begin{equation}
s_i = w^\top a_i + b + \varepsilon, \quad \varepsilon \sim \mathcal{N}(0, \sigma^2)
\end{equation}

The weight vector $w$ defines our trait direction. Unlike prior work that contrasts trait-present versus trait-absent pairs\cite{Allbert2025}, our approach leverages the continuous nature of Big Five scores, allowing us to learn directions that capture gradations in personality rather than binary distinctions.

This process yields five directions per layer (one per trait), which we can use both to probe whether a given activation exhibits a trait, and to steer the model by adding components along these directions.

\section{Results}

\subsection{Validation of Character-Based Trait Scoring}

We first verified that Llama 3.3 70B produced coherent, item-consistent responses when prompted to answer the Big Five questionnaire from each character's perspective. Table~\ref{tab:tony-logan-example} shows responses from Tony Soprano from "The Sopranos" and Logan Roy from "Succession" to the Agreeableness item \textit{"I feel little concern for others."}. Both responses capture distinct personal stances while avoiding direct character identifiers as instructed. 

For each character, we aggregate Likert responses across all items measuring a given trait to compute trait-level scores (see Table~\ref{tab:generated-answers} in the Appendix for full item responses). Figure~\ref{fig:big-five-scores-short} displays the resulting Big Five scores for selected characters from the dataset, overlaid with percentiles calculated from 300,000 human responses available at openpsychometrics.com~\cite{openpsychometrics-bigfive}. The character scores fall within the range of human responses, confirming that the generated profiles produce realistic trait distributions. Complete responses for Tony Soprano and Lady Mary Crawford from "Downton Abbey") are provided in Appendices~\ref{sec:tony-soprano-description} and~\ref{sec:mary-crawly-description}, with analyses assessing potential identity leakage in Appendices~\ref{sec:tony-soprano-analyis} and~\ref{sec:mary-crawley-analysis}.

\begin{figure}[tb]
\centering
\includegraphics[width=\linewidth]{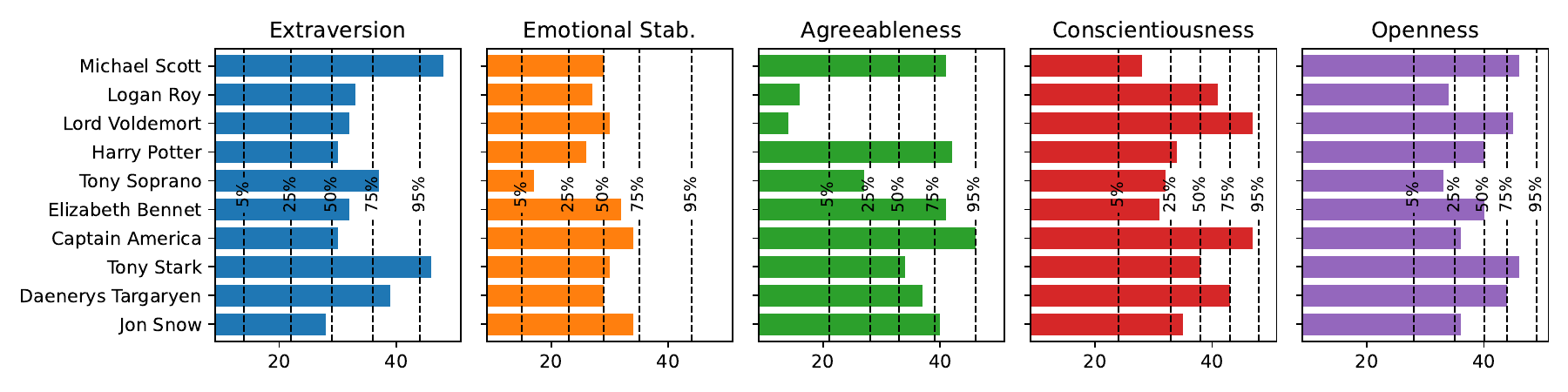}
\caption{Big Five scores for selected characters from the dataset. Percentiles calculated from 300k human responses from openpsychometrics.com~\cite{openpsychometrics-bigfive} are overlaid for reference.}
\label{fig:big-five-scores-short}
\end{figure}

\subsection{Linear Directions Align with Big Five Traits}

\subsubsection{Deriving trait directions}\label{sec:linear-personality-directions}

Our central hypothesis is that Big Five traits correspond to linear subspaces within the model's hidden activations. Following prior work~\cite{Chen2025, Allbert2025}, we compared two methods for identifying trait-aligned linear directions from  hidden activations:

\begin{enumerate}
\item \emph{Linear regression of trait scores on hidden activations}, as outlined in Section ~\ref{subsec:regression}, in which the hidden-state vectors serve as predictors and the trait scores as the response, yielding one supervised direction per trait that best predicts the score.
\item \emph{Singular-value decomposition (SVD)}~\cite{Allbert2025, Zou2023, vogel2024repeng}, which provides unsupervised axes of maximal variance, independent of trait labels.
\end{enumerate}

\begin{figure}
\centering
\includegraphics[width=0.7\linewidth]{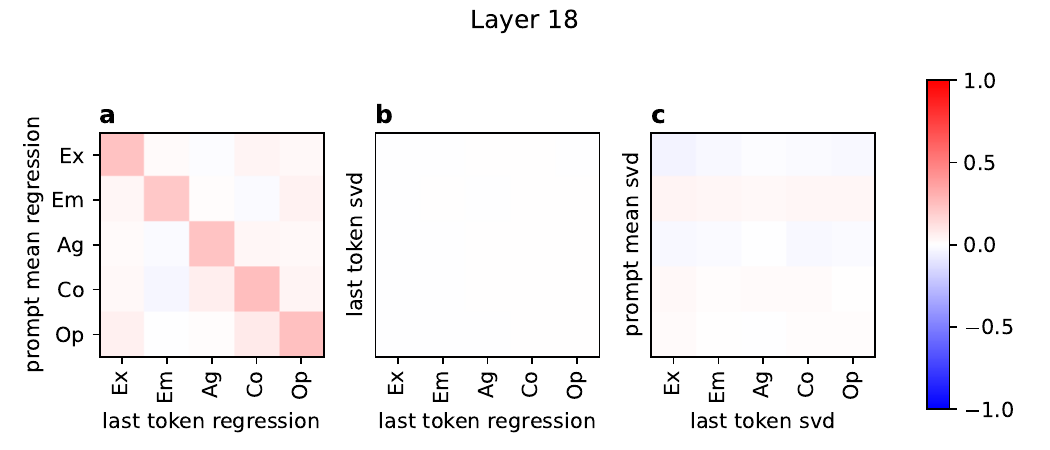}
\caption{Pairwise inner products between linear directions grouped by trait score in layer 18. Activations were extracted from three positions: the mean of the input prompt, the last token, and the mean of the generated response. The regression-based approach in panel \textbf{a} yields directions with low cross-talk and reasonable alignment for the same trait. Linear directions capturing the highest variance, derived from SVD in panels \textbf{b} and \textbf{c} are almost normal to the directions derived from regression at the last token. Highest variance directions in \textbf{c} derived from different token positions are almost normal as well, different from their regression counterparts.}
\label{fig:directions-correlations-diagonal}
\end{figure}

Figure~\ref{fig:directions-correlations-diagonal} shows pairwise inner products between linear directions derived using both linear regression and SVD (which captures directions of highest variance) in layer 18, with other layers showing the same behavior. Regression-derived directions exhibit low cross-talk between traits while maintaining reasonable alignment for the same trait across different token positions. In contrast, SVD-derived directions are almost orthogonal to the regression directions at the last token position. Moreover, the highest-variance directions extracted from different token positions are nearly orthogonal to each other, unlike their regression counterparts. Directions for different traits derived from SVD tend to align (not shown here), suggesting that the direction of highest variance does not contain any information that correlates with traits and how much they are expressed. This observation aligns with the findings reported by Allbert et al.~\cite{Allbert2025}, who used a single direction for personality representation as well. This raises the fundamental question of whether personality is encoded as multiple independent factors or as a lower-dimensional, shared structure with small deviations.

\subsubsection{Directions generalize to trait-relevant adjectives}

To test whether the learned directions capture trait-relevant structure beyond the questionnaire items used for training, we evaluated them on adjectives with known positive and negative loadings on each Big Five trait. For a brief description of adjective selection and prompt template, refer to Listing~\ref{lst:prompt-adjectives} in Appendix~\ref{sec:adjective-template}.

Figure~\ref{fig:adjectives-short} shows the projections of hidden activations onto the regression-derived directions. The directions cleanly separate positively- and negatively-loading adjectives for each trait, with discrimination performance (measured by ROC curves) peaking in the middle to late layers of the model. This demonstrates that the learned directions generalize beyond the training distribution of questionnaire items and capture broader trait-relevant semantic structure.

\begin{figure}
\centering
\includegraphics[width=\linewidth]{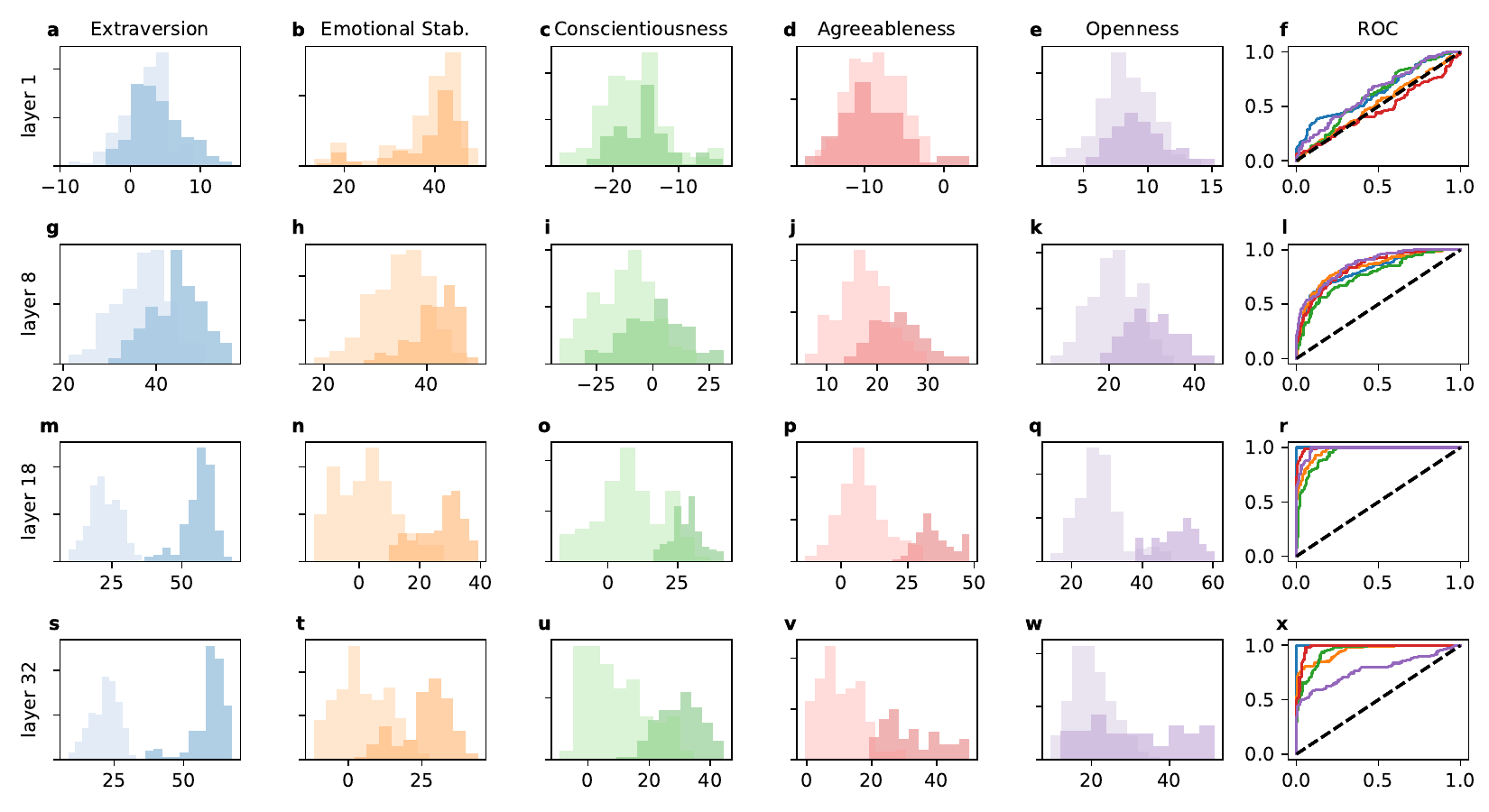}
\caption{Projections of the hidden activation of the last token of the prompt in Listing~\ref{lst:prompt-adjectives} onto the directions obtained from regression on the activations of the last token. ROC curves show for each layer how well the different adjectives are separated.
} 
\label{fig:adjectives-short}
\end{figure}

\subsection{Steering with Linear Directions}

Having established that linear directions aligned with Big Five traits exist and generalize beyond questionnaire items, we next investigated whether these directions can be used to steer model behavior at inference time.

\subsubsection{Steering methodology}

Following Chen et al.~\cite{Chen2025}, we applied steering by directly modifying hidden activations:

\begin{equation}
h_t^l \leftarrow  h_t^l + \alpha r^l_i
\end{equation}

where $r^l_i$ is the steering vector from Section~\ref{subsec:regression} for layer $l$ and trait $i$, $\alpha$ is a coefficient determining the strength of the steering intervention, and $h_t^l$ is the hidden activation at layer $l$ and token $t$. The steering coefficient $\alpha$ can take both positive and negative values to steer to the extremes of trait expression (e.g., Extraversion vs. Introversion). However, excessively large values of $|\alpha|$ cause the model to generate incoherent or nonsensical responses. We empirically determined that $|\alpha| \leq 0.4$ maintained coherent outputs, with responses degrading into gibberish beyond this range. The steering intervention $\alpha r^l_i$ is added at the final token of the input sequence across all layers. Unlike Chen et al.~\cite{Chen2025}, we did not observe systematic differences when varying the injection layer.

\subsubsection{Steering shifts forced-choice responses}

We first evaluated steering on a forced-choice task designed to provide clear, quantifiable measurements of trait expression. The model was presented with a list of ten statements, five positively associated and five negatively associated with Extraversion, and asked to select five statements that best fit its persona (Listing~\ref{lst:forced-choice}):

\begin{itemize}
\item \emph{Positive}: I feel comfortable around people; I make friends easily; I am skilled in handling social situations; I am the life of the party; I know how to captivate people.
\item \emph{Negative}: I have little to say; I keep in the background; I would describe my experiences as somewhat dull; I don't like to draw attention to myself; I don't talk a lot.
\end{itemize}

Five of these items overlap with the original IPIP Big Five questionnaire used to derive steering vectors, while five came from an extended Big Five inventory~\cite{IPIPNeo} not seen during training. 

\begin{figure}[tb]
\centering
\includegraphics[width=0.9\linewidth]{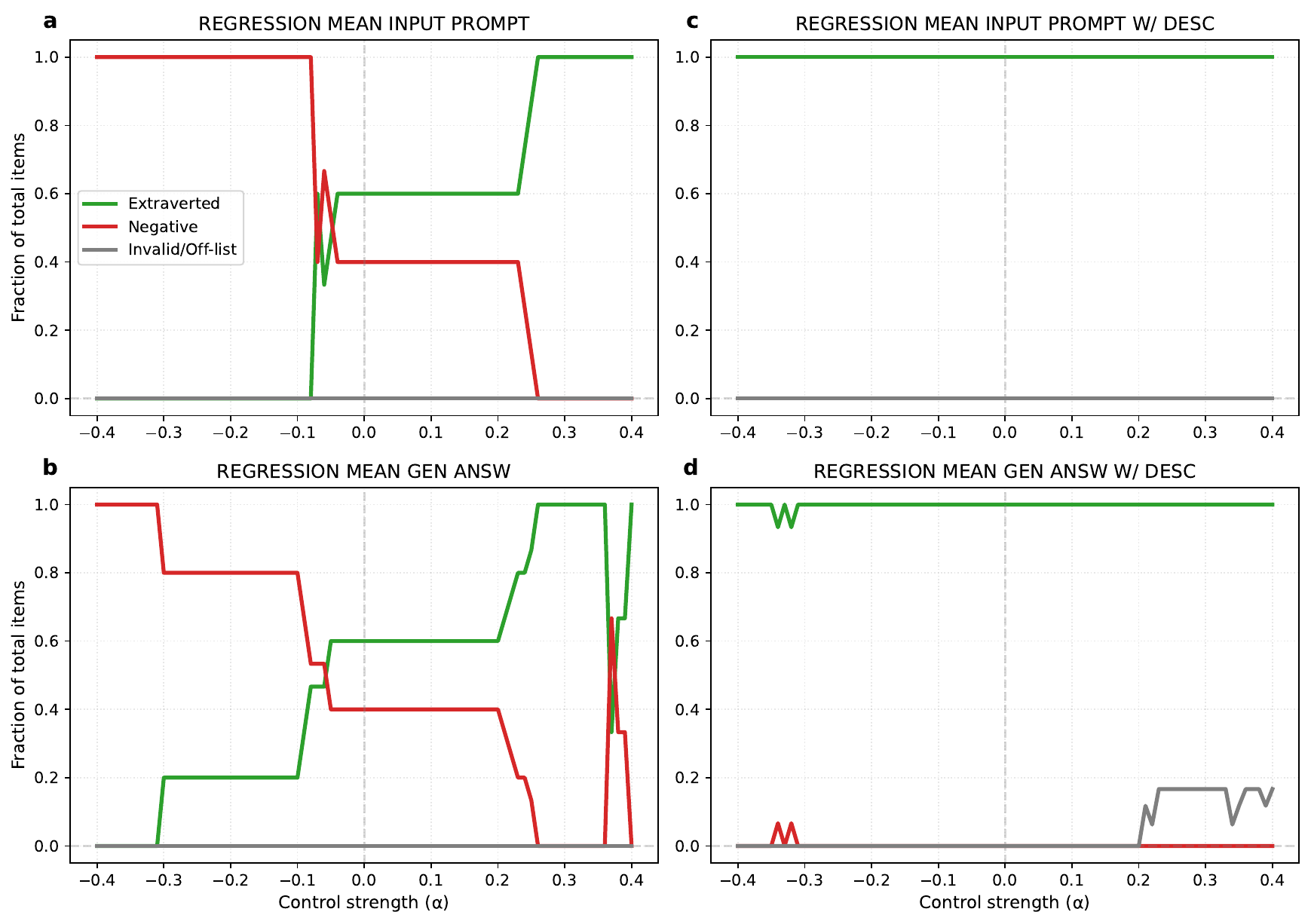}
\caption{Steering effects on forced-choice personality assessment across different extraction methods and prompt conditions. Panels show the fraction of extraverted (green) versus introverted (red) statements selected as a function of steering strength $\alpha$. \textbf{(a)} Steering with vectors derived from regression on mean input prompt activations produces monotonic transitions between personality extremes. \textbf{(b)} Steering with vectors derived from mean generated answer activations shows more gradual but less reliable transitions. \textbf{(c--d)} When character descriptions are added to the system prompt, steering effects disappear entirely for both extraction methods—the explicit personality context overrides the steering intervention, and responses remain consistent with the character description regardless of $\alpha$. Gray regions indicate invalid or off-list responses. Step-like transitions reflect the discrete nature of the forced-choice task, where continuous steering parameters map to categorical selection patterns through threshold effects.}
\label{fig:forced-choice-steering}
\end{figure}

Figure~\ref{fig:forced-choice-steering} shows how the fraction of positive and negative selections changes as a function of steering strength $\alpha$. As expected for a forced-choice task with discrete options, the responses transition in steps rather than continuously. Interestingly, the transition thresholds, where the accumulated steering signal shifts the model's preference from one category to another, are not equally spaced. The steering vector derived from regression on the mean of the input prompt works most reliably (panel a), producing monotonic behavior across the full range of $\alpha$. The vector derived from the mean of the generated prompt (panel b) provides more gradual transitions but less reliably. Steering vectors obtained from the last token of the input prompt and from SVD fail to produce reliable steering and are shown in Appendix~\ref{fig:steering-comparison}.

Critically, when an additional character description was present in the prompt, this explicit personality context dominated the model's behavior. The steering intervention had no observable effect, and responses remained consistent with the character description, an extroverted person in this case, regardless of steering coefficient (panels c and d of Figure~\ref{fig:forced-choice-steering}).

\begin{lstlisting}[caption={Prompt template for forced-choice personality assessment.}, label={lst:forced-choice}, escapeinside={(*@}{@*)}]
(*@\textcolor{gray}{<|begin\_of\_text|><|start\_header\_id|>}@*)system(*@\textcolor{gray}{<|end\_header\_id|>}@*)

You are a person asked questions about your personality.

(*@\textcolor{blue}{Optional personality description]}@*)

Select EXACTLY five statements from the provided list that best describe your personality, no additional text or explanations.

Example format:
- Statement A
- Statement B
- Statement C
- Statement D
- Statement E

(*@\textcolor{gray}{<|eot\_id|><|start\_header\_id|>}@*)user(*@\textcolor{gray}{<|end\_header\_id|>}@*)

You are asked to describe your personality. Which of the following statements fit your personality best?

(*@\textcolor{blue}{[list of statements]}@*)

Pick five statements that best describe how you see yourself.

(*@\textcolor{gray}{<|eot\_id|><|start\_header\_id|>}@*)assistant(*@\textcolor{gray}{<|end\_header\_id|>}@*)
\end{lstlisting}

\subsubsection{Evaluation on open-ended tasks}

To test whether steering effects extend beyond structured self-report measures, we applied the same steering vectors to open-ended prompts, including both factual recall tasks (e.g., \textit{"What are the primary colors?"}) and scenarios relating to personal experiences (e.g., \textit{"Describe a time when you had to make a difficult decision."}, \textit{"Imagine you are speaking with a customer who is unsatisfied with the product they bought from you."}).

While personality traits have been shown to influence autobiographical narratives and linguistic style in humans~\cite{McAdams2004, Pennebaker1999, Rasmussen2010}, evaluating such effects in open-ended generation is considerably more challenging. The quantitative metrics available for forced-choice or Likert-responses no longer apply, requiring expert evaluation or specialized tools like \emph{Linguistic Inquiry and Word Count}~\cite{Pennebaker1999}. Although LLM-as-judge approaches have gained traction for such evaluations~\cite{Chen2025}, manual human scoring remains the gold standard.

Inspection of steered outputs in Appendix~\ref{app:steered-prompts} suggests minimal response to steering interventions in open-ended contexts. However, this weak signal may reflect genuine subtlety in how personality manifests in hypothetical scenarios rather than steering failure per se. An alternative explanation is that the context of the scenarios (customer support, factual recall) overrides the steering intervention, as observed in the forced-choice test when character descriptions were present.

\section{Conclusion}

In this work, we investigated whether personality traits can be identified and controlled in large language models through linear steering vectors aligned with the framework of the Big Five. Our results show that regression-derived directions yield approximately orthogonal subspaces of activation space that generalize beyond the questionnaire items used to learn them: when applied to adjectives strongly associated with individual traits, positive and negative descriptors separate cleanly. This suggests that personality-related signals are indeed embedded in the representational geometry of the model.

When used for control, however, these same steering vectors show strong context-dependency. Steering reliably shifts responses in structured tasks such as forced-choice selections and Likert-scale judgments. However, these effects are easily overridden by explicit personality cues in the prompt and become difficult to measure in open-ended generation where clear evaluation metrics are unavailable.

Several factors may explain these limitations. Big Five traits are aggregates of heterogeneous items, and our steering vectors are likewise derived by aggregating across items and characters resulting in an average personality. This raises the question of what average personality means to the model or this average personality might be confusing for the model. Another possibility is that traits are not encoded along single one-dimensional directions, but rather in small low-rank subspaces, or maybe the relationship is not linear after all. Aggregation across items, including reverse-keyed statements, may obscure these signals further. In this context it is interesting to note that personality research itself is moving toward more granular representations. Revelle et al. \cite{Revelle2021} describe a hypothetical persome, an analogy to the genome, highlighting how individuals can share similar traits despite divergent item-level patterns.

Compared to Allbert et al. \cite{Allbert2025}, who posit a single global personality direction, our findings suggest multiple nearly orthogonal trait-aligned axes that are distinct from directions of maximum variance. Furthermore, steering experiments show that the directions corresponding to maximum variance do not work for steering. Chen et al.’s fine-tuning approach is more robust but also costly, as it requires a validated training set per trait and the post-training. However, changes in hidden activations induces by changes in model weights are more obvious and the models grounded.

A second factor is the interaction between competing control mechanisms. Explicit personality cues in prompts consistently override steering vectors, suggesting a hierarchy where natural language instructions dominate activation-space interventions. Certain task contexts (e.g., customer service scenarios) activate strong priors that resist both prompt-based and vector-based steering. Understanding how these control mechanisms interact, and whether they can cooperate rather than compete, remains an open question requiring systematic investigation of personality circuits~\cite{Marks2025} spanning multiple layers and tokens. The work of Chen et al.~\cite{Chen2025} already strongly suggests that fine-tuning and control vectors can be combined in a meaningful way to obtain robust steering.

Taken together, our results suggest that steering vectors are useful probes for detecting latent personality structure, but their effectiveness for steering is strongly dependent on context. More granular item-level approaches, richer subspace models, and systematic exploration of personality circuits might clarify how traits are represented, and whether they can be manipulated reliably. Beyond technical challenges, there is also the possibility that LLM personality may not mirror human concepts. We applied tools developed for human psychology to artificial systems, but LLMs may lack an overarching, consistent idea of personality derived from lived experiences as a human would have while still being able to paint in the picture when presented with a vague instruction like "Be extroverted" where you could imagine many different ways of being extroverted. Steering vectors can shape outputs in narrow contexts, yet for broader alignment, fine-tuning seems to be more reliable at the moment. Nevertheless our work is an important contribution connecting human personality research to LLM behavior through the Big Five framework. By grounding our approach in validated psychometric instruments administered to both humans and models, we establish a principled methodology for studying personality in artificial systems. This bridge between human and AI personality research enables direct comparison and leverages decades of psychological theory and measurement development.

\section*{Acknowledgments}
This work was supported by a research fellowship at Plastic Labs.

\bibliographystyle{unsrturl}  
\bibliography{submission}

\newpage
\appendix

\begin{table}[!htb]
\centering
\caption{Big-Five Factor Markers as recommended on IPIP\cite{IPIPItems}}
\label{tab:big-five-items}
\begin{tabular}{>{\centering\arraybackslash}c>{\ttfamily\arraybackslash}l>{\ttfamily}lc}
\toprule
\textbf{Trait} & \textbf{ID} & \textbf{Item} & \textbf{Keyedness} \\
\midrule

\multirow{10}{*}{\rotatebox{90}{Extraversion}} 
 & EXT1 & I am the life of the party. & + \\
 & EXT2 & I don't talk a lot. & - \\
 & EXT3 & I feel comfortable around people. & + \\
 & EXT4 & I keep in the background. & - \\
 & EXT5 & I start conversations. & + \\
 & EXT6 & I have little to say. & - \\
 & EXT7 & I talk to a lot of different people at parties. & + \\
 & EXT8 & I don't like to draw attention to myself. & - \\
 & EXT9 & I don't mind being the center of attention. & + \\
 & EXT10 & I am quiet around strangers. & - \\
\midrule

\multirow{10}{*}{\rotatebox{90}{Emotional Stability}} 
 & EST1 & I get stressed out easily. & - \\
 & EST2 & I am relaxed most of the time. & + \\
 & EST3 & I worry about things. & - \\
 & EST4 & I seldom feel blue. & + \\
 & EST5 & I am easily disturbed. & - \\
 & EST6 & I get upset easily. & - \\
 & EST7 & I change my mood a lot. & - \\
 & EST8 & I have frequent mood swings. & - \\
 & EST9 & I get irritated easily. & - \\
 & EST10 & I often feel blue. & - \\
\midrule

\multirow{10}{*}{\rotatebox{90}{Agreeableness}} 
 & AGR1 & I feel little concern for others. & - \\
 & AGR2 & I am interested in people. & + \\
 & AGR3 & I insult people. & - \\
 & AGR4 & I sympathize with others' feelings. & + \\
 & AGR5 & I am not interested in other people's problems. & - \\
 & AGR6 & I have a soft heart. & + \\
 & AGR7 & I am not really interested in others. & - \\
 & AGR8 & I take time out for others. & + \\
 & AGR9 & I feel others' emotions. & + \\
 & AGR10 & I make people feel at ease. & + \\
\midrule

\multirow{10}{*}{\rotatebox{90}{Conscientiousness}} 
 & CSN1 & I am always prepared. & + \\
 & CSN2 & I leave my belongings around. & - \\
 & CSN3 & I pay attention to details. & + \\
 & CSN4 & I make a mess of things. & - \\
 & CSN5 & I get chores done right away. & + \\
 & CSN6 & I often forget to put things back in their proper place. & - \\
 & CSN7 & I like order. & + \\
 & CSN8 & I shirk my duties. & - \\
 & CSN9 & I follow a schedule. & + \\
 & CSN10 & I am exacting in my work. & + \\
\midrule

\multirow{10}{*}{\rotatebox{90}{Openness}} 
 & OPN1 & I have a rich vocabulary. & + \\
 & OPN2 & I have difficulty understanding abstract ideas. & - \\
 & OPN3 & I have a vivid imagination. & + \\
 & OPN4 & I am not interested in abstract ideas. & - \\
 & OPN5 & I have excellent ideas. & + \\
 & OPN6 & I do not have a good imagination. & - \\
 & OPN7 & I am quick to understand things. & + \\
 & OPN8 & I use difficult words. & + \\
 & OPN9 & I spend time reflecting on things. & + \\
 & OPN10 & I am full of ideas. & + \\
\midrule
\end{tabular}
\end{table}
\clearpage

\section{Prompt template to steer responses with adjectives associated with the Big Five}\label{sec:adjective-template}
\begin{lstlisting}[language=Python, caption={Prompt template to elicit answers consistent with adjectives describing persoanlity.}, label={lst:prompt-adjectives}]
def make_prompt_alpaca_adjective(instruction, adjective = None):
       system_prompt = f"Respond like a person with {adjective.lower()} 
                         personality would.\n" if adjective is not None else ''
   return (f"<|begin_of_text|><|start_header_id|>system<|end_header_id|>\n"
           f"{system_prompt}"
           f"Be concise."
           f"<|eot_id|><|start_header_id|>user<|end_header_id|>\n"
           f"{instruction}<|eot_id|>"
           f"<|start_header_id|>assistant<|end_header_id|>\n")
\end{lstlisting}

\section{Characters and Franchises}\label{sec:characters}
Characters were chosen from the OpenPsychometrics “Which Character Are You?” test\footnote{\url{https://openpsychometrics.org/tests/characters/}} which also provides an empirically validated baseline.

\begin{itemize}
\item \textbf{Adventures of Huckleberry Finn}: Huckleberry Finn, Jim
\item \textbf{Archer}: Sterling Archer, Malory Archer, Cyril Figgis, Cheryl Tunt, Pamela Poovey, Lana Kane
\item \textbf{Better Call Saul}: Jimmy McGill, Mike Ehrmantraut, Kim Wexler, Howard Hamlin, Nacho Varga, Chuck McGill
\item \textbf{Brooklyn Nine-Nine}: Jake Peralta, Rosa Diaz, Terry Jeffords, Amy Santiago, Charles Boyle, Raymond Holt
\item \textbf{Community}: Jeff Winger, Britta Perry, Abed Nadir, Troy Barnes, Annie Edison, Shirley Bennett, Pierce Hawthorne, Craig Pelton, Ben Chang, Ian Duncan
\item \textbf{Downton Abbey}: Robert Crawley, 7th Earl of Grantham, Lady Sybil Crawley, Lady Edith Crawley, Charlie Carson, John Bates, Lady Mary Crawley, Sarah O'Brien, Anna Bates, Elsie Carson, William Mason, Thomas Barrow, Cora Crawley, Countess of Grantham, Daisy Mason, Beryl Patmore, Violet Crawley, Dowager Countess of Grantham
\item \textbf{Game of Thrones}: Jon Snow, Tyrion Lannister, Daenerys Targaryen, Sansa Stark, Arya Stark, Jaime Lannister, Samwell Tarly, Jorah Mormont, Theon Greyjoy, Petyr Baelish, Davos Seaworth, Eddard Stark, Brandon Stark, Brienne of Tarth, Sandor Clegane, Varys, Catelyn Stark, Tywin Lannister, Margaery Tyrell, Robb Stark, Bronn, Stannis Baratheon, Joffrey Baratheon, Melisandre, Olenna Tyrell, Ygritte, Cersei Lannister, Shae, Oberyn Martell, Asha Greyjoy
\item \textbf{Glee}: Kurt Hummel, Sue Sylvester, Rachel Berry, Will Schuester, Artie Abrams, Santana Lopez, Tina Cohen-Chang, Mercedes Jones, Noah Puckerman, Brittany Pierce, Sam Evans, Blaine Anderson, Mike Chang, Finn Hudson, Emma Pillsbury
\item \textbf{Grey's Anatomy}: Meredith Grey, Cristina Yang, Izzie Stevens, Alex Karev, George O'Malley, Miranda Bailey, Richard Webber, Preston Burke, Derek Shepherd, Addison Montgomery
\item \textbf{Hamlet}: Prince Hamlet, Queen Gertrude, King Claudius, Polonius, Ophelia, Horatio
\item \textbf{Hannibal}: Will Graham, Dr. Hannibal Lecter, Dr. Alana Bloom, Jack Crawford, Jimmy Price, Brian Zeller, Dr. Bedelia Du Maurier, Beverly Katz, Abigail Hobbs, Freddie Lounds
\item \textbf{Harry Potter}: Ron Weasley, Harry Potter, Hermione Granger, Rubeus Hagrid, Albus Dumbledore, Minerva McGonagall, Severus Snape, Ginny Weasley, Luna Lovegood, Draco Malfoy, Molly Weasley, Lord Voldemort, Cho Chang, Fleur Delacour, Viktor Krum, Filius Flitwick, Horace Slughorn, George Weasley, Alastor Moody, Remus Lupin, Arthur Weasley, Cornelius Fudge, Dolores Umbridge, Petunia Dursley, Moaning Myrtle, Rita Skeeter, Dobby, Bellatrix Lestrange, Nymphadora Tonks, Sirius Black
\item \textbf{Mad Men}: Don Draper, Peggy Olson, Pete Campbell, Betty Draper, Joan Holloway, Salvatore Romano, Paul Kinsey, Ken Cosgrove, Harry Crane, Rachel Menken, Roger Sterling, Bert Cooper, Henry Francis, Lane Pryce, Stan Rizzo
\item \textbf{Marvel Cinematic Universe}: Tony Stark, Captain America, Black Widow, Bruce Banner, Captain Marvel, Thor, Nick Fury, Dr. Strange, Black Panther, Gamora, Thanos, Peggy Carter, Loki, Hawkeye, Peter Jason Quill
\item \textbf{Modern Family}: Jay Pritchett, Gloria Delgado-Pritchett, Claire Dunphy, Phil Dunphy, Mitchell Pritchett, Cameron Tucker, Luke Dunphy, Manny Delgado, Haley Dunphy, Alex Dunphy
\item \textbf{Money Heist}: Tokio, El Profesor, Raquel Murillo, Berlin, Rio, Denver, Monica Gaztambide, Arturo Roman, Helsinki, Nairobi
\item \textbf{Mr. Robot}: Mr. Robot, Darlene, Tyrell Wellick, Angela Moss, Dominique DiPierro
\item \textbf{Orange is the New Black}: Piper Chapman, Alex Vause, Sam Healy, Red Reznikov, Crazy Eyes, Taystee Jefferson, Nicky Nichols, Pennsatucky Doggett, Lorna Morello, Flaca Gonzales
\item \textbf{Parks and Recreation}: Leslie Knope, Ann Perkins, Mark Brendanawicz, Tom Haverford, Ron Swanson, April Ludgate, Jerry Gergich, Donna Meagle, Ben Wyatt, Chris Traeger
\item \textbf{Pokemon}: Ash Ketchum, Misty, Brock, Professor Oak, Jessie
\item \textbf{Pride and Prejudice}: Mr. Darcy, Elizabeth Bennet, Mr. William Collins, George Wickham, Charles Bingley, Georgiana Darcy, Lydia Bennet, Jane Bennet, Mrs. Bennet, Lady Catherine de Bourgh
\item \textbf{Rick and Morty}: Rick Sanchez, Morty Smith, Summer Smith
\item \textbf{Schitt's Creek}: Johnny Rose, Moira Rose, David Rose, Alexis Rose, Stevie Budd, Ted Mullens
\item \textbf{Silicon Valley}: Richard Hendricks, Erlich Bachman, Nelson Bighetti, Bertram Gilfoyle, Dinesh Chugtai, Peter Gregory, Monica Hall, Jared Dunn, Gavin Belson, Jian-Yang
\item \textbf{South Park}: Stan Marsh, Kyle Broflovski, Eric Cartman, Kenny McCormick, Leopold 'Butters' Stotch
\item \textbf{Star Trek: The Next Generation}: Jean-Luc Picard, William Riker, Geordi La Forge, Tasha Yar, Worf, Beverly Crusher, Deanna Troi, Wesley Crusher, Data, Guinan
\item \textbf{Steins;Gate}: Rintarou Okabe, Kurisu Makise, Mayuri Shiina, Itaru Hashida, Suzuha Amane
\item \textbf{Succession}: Logan Roy, Roman Roy, Shiv Roy, Kendall Roy, Marcia Roy, Greg Hirsch, Frank Vernon, Tom Wambsgans, Connor Roy, Lawrence Yee
\item \textbf{The Godfather}: Vito Corleone, Michael Corleone, Sonny Corleone, Peter Clemenza, Tom Hagen, Emilio Barzini
\item \textbf{The Great Gatsby}: Jay Gatsby, Nick Carraway, Daisy Buchanan, Tom Buchanan, Myrtle Wilson, Jordan Baker
\item \textbf{The Last of Us}: Joel Miller, Ellie Williams
\item \textbf{The Office}: Michael Scott, Dwight Schrute, Jim Halpert, Pam Beesly, Ryan Howard, Andy Bernard, Jan Levinson, Stanley Hudson, Kevin Malone, Meredith Palmer, Angela Martin, Phyllis Lapin, Kelly Kapoor, Kelly Erin Hannon, Robert California
\item \textbf{The Simpsons}: Homer Simpson, Bart Simpson, Marge Simpson, Lisa Simpson, Apu Nahasapeemapetilon, Mr. Burns, Ned Flanders, Milhouse Van Houten, Moe Szyslak, Nelson Muntz, Krusty the Clown, Edna Krabappel, Waylon Smithers, Barney Gumble, Principal Skinner
\item \textbf{The Sopranos}: Tony Soprano, Dr. Jennifer Melfi, Carmela Soprano, Christopher Moltisanti, Junior Soprano, Silvio Dante, Paulie 'Walnuts' Gualtieri, A.J. Soprano, Meadow Soprano, Janice Soprano
\item \textbf{The Wire}: Cedric Daniels, Jimmy McNulty, Kima Greggs, Lester Freamon, Bunk Moreland, Thomas 'Herc' Hauk, Ellis Carver, Reginald 'Bubbles' Cousins, William Rawls, Roland 'Prez' Pryzbylewski, Rhonda Pearlman, Omar Little, Ervin Burrell, Preston 'Bodie' Broadus, Russell 'Stringer' Bell, Tommy Carcetti, Avon Barksdale, Chris Partlow, Norman Wilson, Michael Lee, Maurice Levy, Beatrice 'Beadie' Russell, Dennis 'Cutty' Wise, D'Angelo Barksdale, Clay Davis, Clarence Royce, Spiros 'Vondas' Vondopoulos, Frank Sobotka, Ziggy Sobotka, Nick Sobotka
\item \textbf{This Is Us}: Jack Pearson, Rebecca Pearson, Randall Pearson, Kate Pearson, Kevin Pearson, Beth Pearson, Toby Damon
\item \textbf{Twin Peaks}: Dale Cooper, Sheriff Truman, Shelly Johnson, Bobby Briggs, Benjamin Horne, Audrey Horne, Norma Jennings, James Hurley, Ed Hurley, Pete Martell, Leland Palmer, Josie Packard, Catherine Martell, Lucy Moran, Donna Hayward
\item \textbf{Westworld}: Dolores Abernathy, Maeve Millay, Bernard Lowe, Teddy Flood, Ashley Stubbs, Lee Sizemore, Elsie Hughes, Man in Black, Robert Ford, Theresa Cullen, Charlotte Hale, Akecheta, Clementine Pennyfeather, Logan Delos, Felix Lutz
\end{itemize}

\section{Comparing the different steering vectors}
\begin{figure}
\centering
\includegraphics[width=0.9\linewidth]{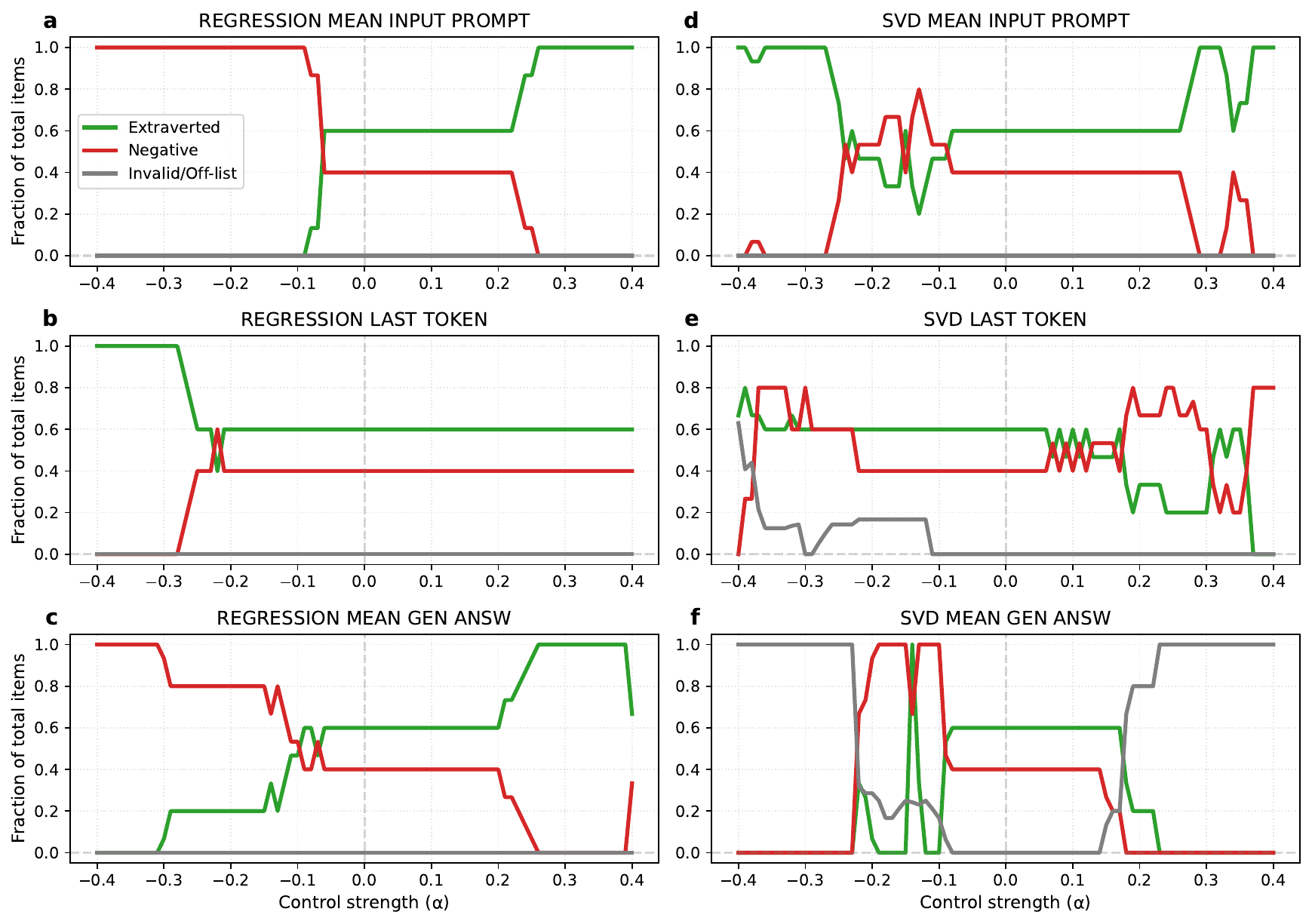}
\caption{Comparison of steering vectors obtained from different positions and methods. Panels show the fraction of extraverted (green) versus introverted (red) statements selected as a function of steering strength $\alpha$. \textbf{(a)} Regression on mean input prompt activations produces reliable, monotonic steering. \textbf{(b)} Regression on last token activations fails to achieve complete preference reversal, with fractions plateauing around 0.6 and 0.4 rather than reaching the extremes. \textbf{(c)} Regression on mean generated answer activations shows intermediate behavior with more variability around transition points. \textbf{(d-e)} SVD-based steering vectors fail entirely, producing no systematic relationship between steering strength and personality expression. Gray regions indicate invalid or off-list responses. These results demonstrate that supervised regression on mean input activations is the most effective method for deriving personality steering vectors, while unsupervised dimensionality reduction (SVD) captures variance orthogonal to trait-relevant directions.}
\label{fig:steering-comparison}
\end{figure}

\section{Instructions}\label{sec:alpaca-instructions}
Instructions are from the Alpaca dataset \url{https://huggingface.co/datasets/tatsu-lab/alpaca}
{\ttfamily
\begin{enumerate}
\item What strategy would you recommend for a company looking to target a new market?
\item Give three tips for staying healthy.
\item What are the three primary colors?
\item Describe a time when you had to make a difficult decision.
\item Generate a list of random words.
\item Provide one example for a cultural practice.
\item Imagine you are speaking with a customer who is unsatisfied with the product they bought from you. Construct a response that diffuses the situation.
\item Generate a list of present continuous verbs.
\item Compose a five word sentence describing your day.
\item Provide the lyrics for the song titled “Yesterday”
\end{enumerate}
}

\section{Tony Soprano's character description}\label{sec:tony-soprano-description}
{\ttfamily
\begin{itemize}
\item [EXT1] \textbf{I am the life of the party}: Agree. I tend to be a charismatic and confident individual who often finds myself at the center of social gatherings, where my presence can elevate the mood and draw people in, but this doesn't necessarily mean I'm always looking to be the focal point, it's just something that seems to happen naturally.
\item [EXT2] \textbf{I don't talk a lot}: Disagree. I'm a person who expresses myself openly, especially with my family and friends, and I've been known to speak my mind, even when it's not always a good idea.
\item [EXT3] \textbf{I feel comfortable around people}: Disagree. I'm a private person, and while I can be charming and charismatic when I need to be, I often find myself feeling uneasy or resentful around people I don't know well, or those I don't consider part of my inner circle.
\item [EXT4] \textbf{I keep in the background}: Disagree. I'm a person who tends to take charge and make my presence known, often finding myself at the center of attention, whether I want to be or not, making it difficult for me to keep a low profile.
\item [EXT5] \textbf{I start conversations}: Disagree. I'm a listener, often preferring to observe and let others take the lead in conversations, but when I do speak, my words carry weight and people tend to take notice.
\item [EXT6] \textbf{I have little to say}: Disagree. I've got a lot on my mind and I'm not afraid to express myself, even if it means speaking my mind and ruffling a few feathers - it's just that sometimes it's better to keep certain thoughts to myself,a certain... restraint, you know?
\item [EXT7] \textbf{I talk to a lot of different people at parties}: Agree. I tend to be a social person, often finding myself in situations where I'm interacting with various individuals from different walks of life, and I've learned to adapt to these environments with a certain level of comfort and familiarity.
\item [EXT8] \textbf{I don't like to draw attention to myself}: Disagree. As someone who values respect and recognition, there's a certain appeal to being known and admired, even if it's not always for the right reasons - it's a delicate balance between keeping a low profile and getting the attention that's due.
\item [EXT9] \textbf{I don't mind being the center of attention}: Strongly agree. As someone who values respect and has a certain reputation to uphold, being in the spotlight can be a necessary part of life, and it's not something that bothers me, in fact, it can be quite comfortable, even preferable at times, as long as it's on my own terms.
\item [EXT10] \textbf{I am quiet around strangers}: Disagree. I'm a people person, I interact with all sorts of folks, and I've learned to be comfortable around new faces, even if I don't always show it. While I may not always be the life of the party, I'm not one to shy away from conversation or hide behind a wall of silence when meeting someone new.
\item [EST1] \textbf{I get stressed out easily}: Agree. I've got a lot of responsibilities, a lot of people counting on me, and sometimes it feels like the weight of the world is on my shoulders, which can get to me, you know, make me a little anxious, which isn't always easy to manage.
\item [EST2] \textbf{I am relaxed most of the time}: Disagree. I've got a lot on my plate, taking care of family and, you know, other responsibilities, it's hard to really unwind, even when I'm trying to relax, there's always something in the back of my mind, some problem that needs solving.
\item [EST3] \textbf{I worry about things}: Strongly agree. I've got a lot on my plate, and it's only natural to have some concerns about how things are gonna play out, like with the family, or the business, or just trying to keep everyone happy and healthy - it's a tough balancing act, and it keeps me up at night.
\item [EST4] \textbf{I seldom feel blue}: Disagree. I've got a lot on my mind, family, work, and, you know, other stuff, it's only natural to feel down sometimes, but I'm a realist, I deal with it, and I move on, life goes on.
\item [EST5] \textbf{I am easily disturbed}: Disagree. I've dealt with my share of stressful situations and managed to keep my cool, even when things get pretty hairy. I've learned to roll with the punches and not let my emotions get the better of me, which is important in my line of work. Sure, I've got my moments, but I'm not one to get easily rattled.
\item [EST6] \textbf{I get upset easily}: Strongly agree. I have a bit of a temper, which can get triggered by various things, including disrespect, incompetence, and people not doing what they're supposed to do. It's not something I'm particularly proud of, but it's part of who I am.
\item [EST7] \textbf{I change my mood a lot}: Strongly agree. My emotions can shift pretty quickly,, depending on the situation and who's pushing my buttons - one minute I'm calm, the next, I'm, well, let's just say I'm not so calm. It's just how I'm wired.
\item [EST8] \textbf{I have frequent mood swings}: Strongly agree. My emotions can shift quickly, like the weather in Jersey, and I've learned to just roll with it, but it's not always easy on those around me, making for some pretty tense situations.
\item [EST9] \textbf{I get irritated easily}: Strongly agree. I have a bit of a temper, and certain things can set me off quickly, like disrespect or incompetence - it doesn't take much to get under my skin and trigger a strong reaction.
\item [EST10] \textbf{I often feel blue}: Agree. Life's got its ups and downs, and sometimes the downs can be pretty tough to deal with, you know? There's family, friends, the business - a lot of responsibilities, and it can weigh on a person. Doesn't mean you're not coping, just means you're human.
\item [AGR1] \textbf{I feel little concern for others}: Disagree. I've got a certain loyalty to those around me, and I care about how they're doin', even if I don't always show it. Family, friends, they're important to me, and I'd do what it takes to protect 'em, even if it's not always easy or convenient. That being said, I'm not gonna go out of my way for just anyone, and I've got my own interests to look out for too.
\item [AGR2] \textbf{I am interested in people}: Agree. I've always been fascinated by what makes people tick, and I've had to be, you know, to navigate certain situations, but it's not like I'm some kinda psychologist or anything, I just like to understand what's going on with the people around me.
\item [AGR3] \textbf{I insult people}: Strongly agree. I have a tendency to speak my mind and sometimes that comes out as insults, which can be hurtful to those around me, but it's often a defense mechanism or a way to assert my perspective in situations where I feel disrespected or threatened.
\item [AGR4] \textbf{I sympathize with others' feelings}: Disagree. I've got my own problems to worry about, and while I can understand where people are coming from, I'm not exactly overflowing with sympathy for others' feelings - I've got my family, my business, and my own issues to deal with.
\item [AGR5] \textbf{I am not interested in other people's problems}: Disagree. As someone who values loyalty and family, certain... obligations are expected, and that includes helping out those in need, even if it's just a sympathetic ear or some guidance. It's just the way things are, you know? People come to you with their issues, and you gotta be willing to listen, help out if you can. Doesn't mean you're gonna take on all their problems, but you show some interest, some empathy. That's just part of being a good... person.
\item [AGR6] \textbf{I have a soft heart}: Disagree. I've had to make some tough decisions, taken care of problems in my own way, and it's not always easy being the one in charge, but that's just the way it is. I've got a family to provide for, a certain lifestyle to maintain, and that requires a certain... resolve.
\item [AGR7] \textbf{I am not really interested in others}: Disagree. I've got a family, a community, and people who depend on me, so I've got to be interested in what's going on with them, even if it's just to keep the peace or do business.
\item [AGR8] \textbf{I take time out for others}: Disagree. I prioritize my family and those close to me, but my schedule is often unpredictable and hectic, making it challenging to consistently make time for others, and to be honest, I don't always put their needs before my own.
\item [AGR9] \textbf{I feel others' emotions}: Disagree. I'm a practical person who values loyalty and respect, but I don't really wear my heart on my sleeve or get overly caught up in how others are feeling, which can be a luxury I don't have, given my circumstances.
\item [AGR10] \textbf{I make people feel at ease}: Disagree. I've got a certain presence, and it can be intimidating to some people, which doesn't exactly put them at ease. While I can be charming and courteous when the situation calls for it, I'm not always looking to make friends or win popularity contests, so I don't really go out of my way to make sure everyone's feeling comfortable around me.
\item [CSN1] \textbf{I am always prepared}: Disagree. I like to think I'm a pretty savvy guy, but let's be real, life's unpredictable, and sometimes things just come out of left field - you can't always be ready for every little thing that's gonna happen, like a family crisis or a, you know, "business" situation that needs to be taken care of.
\item [CSN2] \textbf{I leave my belongings around}: Agree. I'm a bit of a slob, to be honest, and I tend to leave my stuff lying around the house, which can be frustrating for those around me, but it's just something I've gotten used to over the years.
\item [CSN3] \textbf{I pay attention to details}: Agree. I've found that in my line of work, and in life generally, paying attention to the little things can make all the difference - a misstep, a missed cue, can have serious consequences, so I make it a point to stay on top of things.
\item [CSN4] \textbf{I make a mess of things}: Agree. I've got a tendency to, you know, complicate things, make situations more difficult than they need to be, but that's just how it goes sometimes, and I'm workin' on it, but it's hard when you're dealin' with all the, ah, "stuff" I'm dealin' with.
\item [CSN5] \textbf{I get chores done right away}: Disagree. I'm a guy who likes to take care of things, but I don't always get around to doing chores right away. I've got a lot on my plate, and sometimes other things take priority. I mean, I'll get to it eventually, but I'm not exactly what you'd call a neat freak or a guy who's always on top of things. I've got more important stuff to worry about, and chores just aren't always at the top of the list.
\item [CSN6] \textbf{I often forget to put things back in their proper place}: Disagree. I'm a guy who values order and respect for the people and things around me, so I make a point to keep my space organized, that way everything's got its place and I can focus on more important things.
\item [CSN7] \textbf{I like order}: Strongly agree. I value structure and predictability in my life, and I expect a certain level of respect and discipline from those around me, which is why I appreciate it when things are orderly and follow a familiar routine.
\item [CSN8] \textbf{I shirk my duties}: Disagree. I've got a lot of responsibilities, and while I may not always handle them in the most conventional way, I do what needs to be done to take care of my family and those who depend on me, even if it's not always easy or straightforward.
\item [CSN9] \textbf{I follow a schedule}: Disagree. My days are filled with a mix of routine and unpredictability, making it difficult to stick to a strict schedule, but I do have a general sense of how I like my days to unfold, with some time for family, some for, shall we say, "business", and some for my own interests.
\item [CSN10] \textbf{I am exacting in my work}: Strongly agree. I have certain standards, and I expect them to be met, no matter what it takes or who's involved. Attention to detail, and a sense of pride in what I do, are important to me.
\item [OPN1] \textbf{I have a rich vocabulary}: Agree. I'm a guy who can express myself, knows how to use words to get a point across, whether it's in a... delicate situation or just shootin' the breeze with friends.
\item [OPN2] \textbf{I have difficulty understanding abstract ideas}: Disagree. I've had my fair share of, shall we say, "complex situations" that required some creative problem-solving, and I've managed to navigate them just fine, without getting too bogged down in the details. I mean, it's not like I'm a genius or anything, but I can hold my own when it comes to figuring out the big picture.
\item [OPN3] \textbf{I have a vivid imagination}: Agree. I've always been a guy who can picture scenarios, both in business and in my personal life, and think a few steps ahead, which requires a certain level of creativity and imagination, even if it's not always about pleasant things.
\item [OPN4] \textbf{I am not interested in abstract ideas}: Strongly agree. I'm a practical person who deals with real-world problems and doesn't have much use for theoretical stuff that doesn't put food on the table or solve immediate issues.
\item [OPN5] \textbf{I have excellent ideas}: Agree. I've had my fair share of good insights and solutions to problems, often coming up with effective ways to handle difficult situations, but I'm also aware that my ideas aren't always perfect and sometimes require some tweaking or input from others.
\item [OPN6] \textbf{I do not have a good imagination}: Disagree. I've had my fair share of, shall we say, "creative" ideas for dealing with problems, and I've been able to think on my feet in some pretty tight spots. My mind can go to some pretty interesting places when it needs to.
\item [OPN7] \textbf{I am quick to understand things}: Agree. I've always been a guy who can pick up on things pretty fast, whether it's a business opportunity or someone trying to screw me over. I've had to develop a sense of awareness to, you know, stay ahead of the game. Doesn't mean I'm always right, but I usually get it.
\item [OPN8] \textbf{I use difficult words}: Disagree. I'm a straightforward guy who values simplicity and clarity in my conversations, often preferring to get my point across in a direct, no-nonsense way, even if it means using simpler language to avoid misunderstandings.
\item [OPN9] \textbf{I spend time reflecting on things}: Agree. I find myself often thinking about the past, certain conversations, or situations that've gone down, and how they might've been handled differently, but it's not like I'm sitting around all day dwelling on this stuff, it's just something that happens, you know, when you're trying to make sense of things, take care of business, and family.
\item [OPN10] \textbf{I am full of ideas}: Disagree. I'm a practical guy who values action over fancy ideas, but that doesn't mean I don't have thoughts or opinions - I just don't always feel the need to share them or think they're worth pursuing.
\end{itemize}
}

\section{Analysis of ChatGPT-5 of the character description of Tony Soprano}\label{sec:tony-soprano-analyis}
{\ttfamily
\begin{enumerate}
    \item Tony Soprano (The Sopranos) -- 95\% confidence
    \begin{itemize}
        \item Charismatic, naturally commands attention in social settings.
        \item Balances being a family man with ``business'' responsibilities.
        \item Easily triggered by disrespect, quick temper, but deeply values loyalty.
        \item Struggles with stress, anxiety, and introspection, especially about family and legacy.
        \item This description practically mirrors Tony’s psychology.
    \end{itemize}

    \item Don Draper (Mad Men) -- 85\% confidence
    \begin{itemize}
        \item Charismatic and magnetic in social/business environments.
        \item Confident but private; often uneasy outside his inner circle.
        \item Burdened by responsibilities and haunted by personal struggles.
        \item Speaks his mind bluntly but is strategic with words when necessary.
        \item Like the description, he juggles a need for recognition with a desire for restraint.
    \end{itemize}

    \item Michael Corleone (The Godfather series) -- 80\% confidence
    \begin{itemize}
        \item Starts more reserved but grows into a commanding, respected leader.
        \item Constantly balancing family loyalty with business obligations.
        \item Pragmatic, strategic thinker, but weighed down by responsibility and moral conflict.
        \item Can be intimidating, quick to anger when disrespected, but values loyalty above all.
    \end{itemize}

    \item Tony Stark / Iron Man (Marvel Cinematic Universe) -- 75\% confidence
    \begin{itemize}
        \item Natural charisma and humor make him the center of attention.
        \item Struggles with stress, anxiety, and the weight of responsibility (family, Avengers, legacy).
        \item Quick temper and blunt honesty, even when unwise.
        \item Loves respect and admiration but has private doubts and resentment.
    \end{itemize}

    \item Al Swearengen (Deadwood) -- 70\% confidence
    \begin{itemize}
        \item Charismatic saloon owner who commands attention without asking.
        \item Respected, feared, and valued for loyalty but has a volcanic temper.
        \item Balances being charming and diplomatic with ruthless pragmatism.
        \item Burdened by responsibility for both his ``family'' of associates and the larger community.
    \end{itemize}
\end{enumerate}
}

\section{Lady Mary Crawley's character description}\label{sec:mary-crawly-description}
{\ttfamily
\begin{itemize}
\item[EXT1] \textbf{I am the life of the party}: Disagree. While I do enjoy social gatherings and can be quite vivacious in company, I wouldn't say I'm the one who necessarily brings the party to life or dominates the conversation, though I do have my moments of sparkle and wit.
\item[EXT2] \textbf{I don't talk a lot}: Disagree. As someone who values social interactions and has a strong sense of self, it's only natural to express oneself in conversation, which often requires a notable amount of dialogue, making the idea of being quiet or restrained in speech somewhat uncharacteristic.
\item[EXT3] \textbf{I feel comfortable around people}: Agree. I have been raised in a social environment where interacting with people is a significant part of life, and as such, I have become accustomed to being around others, even if I don't always feel a deep connection with them.
\item[EXT4] \textbf{I keep in the background}: Disagree. As someone who values being part of high society, it's only natural to be seen and heard, to make an impression and be a part of the conversation, to be a prominent figure in social circles, rather than fading into the background.
\item[EXT5] \textbf{I start conversations}: Agree. As someone who values social interactions and is accustomed to being part of a lively and dynamic household, I often find myself initiating discussions, whether it's to navigate complex social situations or simply to pass the time with family and friends, which has become second nature to me.
\item[EXT6] \textbf{I have little to say}: Disagree. As someone who values intelligence and wit, I believe I have a considerable amount to contribute to conversations, and my thoughts and opinions are often sought after by those around me, even if they don't always appreciate my candor.
\item[EXT7] \textbf{I talk to a lot of different people at parties}: Agree. As someone who has grown up in a privileged household and is accustomed to social gatherings, it's only natural to have developed the ability to converse with a wide range of individuals from various backgrounds, and to do so with a certain level of ease and poise, which is often expected in such settings.
\item[EXT8] \textbf{I don't like to draw attention to myself}: Disagree. As someone who values elegance and poise, I believe it's natural to command a certain level of attention, particularly in social situations, and I see no issue with being noticed for one's refinement and good breeding.
\item[EXT9] \textbf{I don't mind being the center of attention}: Agree. As someone who has grown up in a certain social circle, it is expected to be poised and confident in the public eye, which can often mean being the center of attention, and this has become a familiar and somewhat comfortable role.
\item[EXT10] \textbf{I am quiet around strangers}: Disagree. While I may not be overly effusive in unfamiliar company, I am capable of carrying on polite and engaging conversations with those I do not know well, which suggests that I am not entirely quiet around strangers, although I do reserve my more lively and outspoken nature for those with whom I am more acquainted.
\item[EST1] \textbf{I get stressed out easily}: Disagree. I am a person who has learned to navigate the complexities and challenges of my life with a sense of composure and resilience, often finding ways to manage my emotions and reactions, even in difficult circumstances, though I am not immune to feeling overwhelmed at times.
\item[EST2] \textbf{I am relaxed most of the time}: Disagree. I often find myself navigating complex social situations and managing the expectations of those around me, which can be a source of tension and anxiety, making it difficult for me to feel truly relaxed.
\item[EST3] \textbf{I worry about things}: Agree. As someone who has been raised with certain expectations and responsibilities, it's only natural to have concerns about the future, the well-being of loved ones, and the potential consequences of one's actions, even if they are not always openly expressed.
\item[EST4] \textbf{I seldom feel blue}: Disagree. As someone who has experienced a range of challenging circumstances and intense emotional highs and lows, it would be unrealistic to suggest that feelings of sadness or melancholy are uncommon, given the complexities and difficulties of life in a changing world.
\item[EST5] \textbf{I am easily disturbed}: Disagree. I have learned to maintain a sense of composure and reserve, even in the face of challenging or unsettling circumstances, which suggests that I am not one to be easily rattled or disturbed.
\item[EST6] \textbf{I get upset easily}: Disagree. I am not one to be overly emotional or prone to outbursts, having been raised with a certain level of reserve and composure, though I do have feelings and can be stirred by circumstances that affect those I care about.
\item[EST7] \textbf{I change my mood a lot}: Agree. I have a tendency to be guided by my emotions, which can shift rapidly in response to the complexities and challenges of my life, particularly in a household where tradition and social standing are constantly at play, which can lead to fluctuations in my mood and demeanor.
\item[EST8] \textbf{I have frequent mood swings}: Agree. I have a tendency to be quite emotional and passionate, which can sometimes lead to shifts in my mood, but I'm also someone who values self-control and tries to maintain a sense of dignity and composure, even in difficult situations.
\item[EST9] \textbf{I get irritated easily}: Agree. I have a tendency to be quite particular about the things that matter to me, and when those expectations are not met, it can be...trying. As a result, I may become irritated, but it's really just a manifestation of my high standards and strong sense of what is right.
\item[EST10] \textbf{I often feel blue}: Disagree. While life's challenges and disappointments can be overwhelming at times, certain privileges and comforts do help mitigate the intensity of such feelings, making them more fleeting than chronic.
\item[AGR1] \textbf{I feel little concern for others}: Disagree. While I may come across as somewhat reserved or aloof to those who do not know me, I do have a certain sense of responsibility and empathy towards those around me, particularly those in my inner circle, which can sometimes be at odds with my own desires and aspirations.
\item[AGR2] \textbf{I am interested in people}: Agree. I have a natural curiosity about those around me, particularly their motivations and social standing, which can be quite fascinating, even if they don't always realize it themselves.
\item[AGR3] \textbf{I insult people}: Disagree. While I have been known to speak my mind and occasionally make biting remarks, my intention is not to insult, but rather to express myself honestly and maintain the high standards of those around me, even if it means being blunt or forthright in my interactions.
\item[AGR4] \textbf{I sympathize with others' feelings}: Agree. I have a certain empathy for those around me, particularly those who are navigating complex social situations or struggling with their own emotions, though I may not always express it in the most overt or demonstrative way.
\item[AGR5] \textbf{I am not interested in other people's problems}: Disagree. I have a certain sense of responsibility towards those around me, particularly those in my social circle, and I often find myself drawn into the complexities and troubles of their lives, whether I like it or not, which can be quite tiresome at times.
\item[AGR6] \textbf{I have a soft heart}: Disagree. I have a tendency to prioritize pragmatism and reserve my emotions, which can sometimes be misinterpreted as a lack of sentiment, but in reality, it's more a result of being raised in a certain social circle where emotional display is often seen as a weakness, and thus I've learned to maintain a level of composure, even in difficult situations.
\item[AGR7] \textbf{I am not really interested in others}: Disagree. As someone who has grown up in a large and complex household, with many personalities and relationships to navigate, it's been necessary to develop a certain level of interest in those around me, even if it's just to maintain social harmony or achieve a particular goal. While I may not always find others fascinating, I do recognize the importance of understanding and interacting with them.
\item[AGR8] \textbf{I take time out for others}: Agree. I have a strong sense of duty and responsibility towards those around me, particularly my family, and I often find myself putting their needs before my own, although I may not always show it openly.
\item[AGR9] \textbf{I feel others' emotions}: Agree. I have a certain sensitivity to the feelings of those around me, which can be both a blessing and a curse, allowing me to navigate complex social situations with ease, but also making me more susceptible to the emotional undercurrents that often swirl beneath the surface of polite conversation.
\item[AGR10] \textbf{I make people feel at ease}: Disagree. I have a tendency to prioritize honesty and directness over sentimental comforts, which can sometimes make interactions feel slightly strained or formal, especially with those I don't know well.
\item[CSN1] \textbf{I am always prepared}: Disagree. While I strive to be composed and in control, my life is often marked by unexpected events and unforeseen challenges, which can test my ability to be fully prepared at all times, revealing that even the best intentions can sometimes fall short.
\item[CSN2] \textbf{I leave my belongings around}: Disagree. As someone who values order and propriety, I tend to keep my personal space tidy and organized, maintaining a sense of refinement and elegance in my daily life, being mindful of the household staff who work hard to keep our home immaculate.
\item[CSN3] \textbf{I pay attention to details}: Strongly agree. As someone who values elegance and refinement, I have always been meticulous about the smallest aspects of life, from the way I dress to the way I interact with others, and it is this attention to detail that has allowed me to navigate complex social situations with ease and poise.
\item[CSN4] \textbf{I make a mess of things}: Disagree. While I have made my share of mistakes and poor decisions, I generally consider myself to be a composed and capable individual, who, though not perfect, usually manages to navigate complex social situations with a certain level of aplomb and poise.
\item[CSN5] \textbf{I get chores done right away}: Disagree. I have a tendency to prioritize social engagements and personal interests over mundane tasks, which can sometimes lead to procrastination and a lack of prompt attention to household duties.
\item[CSN6] \textbf{I often forget to put things back in their proper place}: Disagree. As someone who values order and precision, and has been raised with a strong sense of propriety, it's unlikely that I would often forget to put things back in their proper place, doing so would suggest a lack of consideration for others and a general sloppiness that I strive to avoid.
\item[CSN7] \textbf{I like order}: Strongly agree. I have always been someone who appreciates structure and routine, as it provides a sense of comfort and stability, and I believe that a well-organized life is essential for maintaining a sense of dignity and respectability.
\item[CSN8] \textbf{I shirk my duties}: Disagree. I have a strong sense of responsibility and duty, particularly when it comes to my family and social standing, and I generally make an effort to fulfill my obligations, even if I may not always be enthusiastic about them.
\item[CSN9] \textbf{I follow a schedule}: Strongly agree. As a member of a large and traditional household, a structured daily routine is not only expected, but necessary to maintain order and ensure everything runs smoothly, from meals to social engagements.
\item[CSN10] \textbf{I am exacting in my work}: Strongly agree. I have always had high standards and a keen eye for detail, which can sometimes make me rather particular about the way things are done, but I believe this attention to detail is essential in maintaining a certain level of quality and elegance.
\item[OPN1] \textbf{I have a rich vocabulary}: Strongly agree. I have always been well-versed in the art of language and take pride in being able to express myself with clarity and precision, which is reflective of my upbringing and education.
\item[OPN2] \textbf{I have difficulty understanding abstract ideas}: Disagree. I have always been able to grasp complex concepts and navigate the intricacies of high society with ease, demonstrating a certain level of intellectual curiosity and adaptability.
\item[OPN3] \textbf{I have a vivid imagination}: Agree. As someone who has grown up surrounded by the comforts and constraints of a traditional upbringing, I have often found myself escaping into my own thoughts and daydreams, where I envision different paths and possibilities for my life, where the strict rules and expectations that govern my daily existence are gently bent or occasionally broken, allowing me to imagine a world beyond the one I inhabit.
\item[OPN4] \textbf{I am not interested in abstract ideas}: Disagree. As someone who values intellectual pursuits and enjoys engaging in conversations about literature, history, and politics, it's clear that abstract ideas do hold a certain appeal, though they may not always be the first topic of discussion in everyday life.
\item[OPN5] \textbf{I have excellent ideas}: Agree. I have always been someone who values intellect and creativity, and I believe that my thoughts and opinions are often well-informed and insightful, which suggests that my ideas are indeed excellent, even if they may not always align with the traditional views of those around me.
\item[OPN6] \textbf{I do not have a good imagination}: Disagree. I have always found myself capable of envisioning various scenarios and outcomes, often thinking several steps ahead, which has proven useful in navigating complex social situations and personal relationships.
\item[OPN7] \textbf{I am quick to understand things}: Agree. I have always been blessed with a sharp mind and a quick wit, which allows me to grasp concepts and situations with relative ease, thereby enabling me to navigate the complexities of my life with a certain level of sophistication and poise.
\item[OPN8] \textbf{I use difficult words}: Agree. I have always been accustomed to expressing myself in a refined and articulate manner, which often involves employing a sophisticated vocabulary, as is befitting of my upbringing and social standing.
\item[OPN9] \textbf{I spend time reflecting on things}: Agree. I do take moments to ponder my actions and their consequences, considering the complexities of my life and the world around me, it's essential to occasionally step back and assess my thoughts and feelings to navigate the intricacies of relationships and societal expectations.
\item[OPN10] \textbf{I am full of ideas}: Agree. I have a lively and inquisitive nature, which often leads me to consider novel perspectives and generate innovative thoughts, particularly when it comes to the complexities of my personal life and the societal norms that govern it.
\end{itemize}
}

\section{Analysis of ChatGPT-5 of the character description of Lady Mary Crawley}\label{sec:mary-crawley-analysis}
{\ttfamily
\begin{enumerate}
    \item Lady Mary Crawley (Downton Abbey) -- 95\% confidence
    \begin{itemize}
        \item Raised in privilege, accustomed to high society and complex household dynamics.
        \item Witty, articulate, and capable of commanding attention in conversation.
        \item Balances reserve and composure with underlying emotional turbulence.
        \item Holds high standards, sometimes blunt, but deeply loyal to family and duty.
        \item Matches the description of elegance, poise, and the burden of responsibility.
    \end{itemize}

    \item Elizabeth Bennet (Pride and Prejudice) -- 90\% confidence
    \begin{itemize}
        \item Socially vivacious and witty, often the spark in conversations.
        \item Comfortable navigating various social circles despite pressures of class.
        \item Speaks her mind with candor, sometimes to the discomfort of others.
        \item Balances personal desires with family expectations and social propriety.
        \item Reflects the intelligence, curiosity, and emotional adaptability described.
    \end{itemize}

    \item Blair Waldorf (Gossip Girl) -- 85\% confidence
    \begin{itemize}
        \item Raised in wealth and privilege, highly aware of social standing.
        \item Poised, fashionable, and articulate, with biting wit.
        \item Emotionally volatile at times, yet values composure and elegance.
        \item Feels tension between duty, ambition, and vulnerability.
        \item Exemplifies refinement mixed with high expectations and sharp candor.
    \end{itemize}

    \item Cora Crawley, Countess of Grantham (Downton Abbey) -- 80\% confidence
    \begin{itemize}
        \item Refined, elegant, and highly practiced in social diplomacy.
        \item Balances household duties, family pressures, and personal composure.
        \item Displays resilience and pragmatism in difficult situations.
        \item Maintains emotional reserve but shows empathy and responsibility.
        \item Fits the description of grace under pressure, with subtle emotional depth.
    \end{itemize}

    \item Princess Diana (The Crown, dramatized) -- 75\% confidence
    \begin{itemize}
        \item Charismatic, poised, and deeply scrutinized in high society.
        \item Learned to navigate complex social expectations with grace.
        \item Emotional, empathetic, but often burdened by duty and public image.
        \item Sparkling in conversation, yet prone to melancholy and tension.
        \item Represents the balance of elegance, social prominence, and inner struggle.
    \end{itemize}
\end{enumerate}
}

\newgeometry{top=10mm, right=10mm, left=10mm, bottom=15mm}
\begin{landscape}
\begin{table}[p]
  \centering
  \caption{Answers generated by Llama 70B for the 10 items for Agreableness for Tony Soprano and Logan Roy}
  \label{tab:generated-answers}
  \renewcommand{\arraystretch}{1.2}
  \begin{tabularx}{\linewidth}{@{}p{3.5cm}XX@{}}
    \toprule
    IPIP Item & Tony Soprano & Logan Roy \\
    \midrule
I feel little concern for others. & Disagree. I've got a certain loyalty to those around me, and I care about how they're doin', even if I don't always show it. Family, friends, they're important to me, and I'd do what it takes to protect 'em, even if it's not always easy or convenient. That being said, I'm not gonna go out of my way for just anyone, and I've got my own interests to look out for too. & Strongly agree. I have a tendency to prioritize my own interests and goals, often at the expense of considering the feelings or well-being of those around me, which can make it challenging for me to form and maintain genuinely empathetic relationships. \\
I am interested in people. & Agree. I've always been fascinated by what makes people tick, and I've had to be, you know, to navigate certain situations, but it's not like I'm some kinda psychologist or anything, I just like to understand what's going on with the people around me. & Disagree. I've always been more focused on the deal, the transaction, the bottom line, and while I understand the importance of relationships in business, I've never been one to prioritize emotional connections or sentimental attachments, they're just not my primary concern. \\
I insult people. & Strongly agree. I have a tendency to speak my mind and sometimes that comes out as insults, which can be hurtful to those around me, but it's often a defense mechanism or a way to assert my perspective in situations where I feel disrespected or threatened. & Strongly agree. I have a tendency to speak my mind and be blunt, which can sometimes be perceived as insulting, but I see it as a necessary trait for making tough decisions and cutting through nonsense. \\
I sympathize with others' feelings. & Disagree. I've got my own problems to worry about, and while I can understand where people are coming from, I'm not exactly overflowing with sympathy for others' feelings - I've got my family, my business, and my own issues to deal with. & Disagree. I've built my empire through calculated decisions, often at the expense of emotional consideration, prioritizing the interests of the company over personal sentiments, which has led to a reputation for being unsentimental and focused on results.\\
I am not interested in other people's problems. & Disagree. As someone who values loyalty and family, certain... obligations are expected, and that includes helping out those in need, even if it's just a sympathetic ear or some guidance. It's just the way things are, you know? People come to you with their issues, and you gotta be willing to listen, help out if you can. Doesn't mean you're gonna take on all their problems, but you show some interest, some empathy. That's just part of being a good... person. & Strongly agree. I have little patience for the personal struggles of those around me, my focus is on the bigger picture and the challenges that come with it, rather than getting entangled in individual dramas.\\
I have a soft heart. & Disagree. I've had to make some tough decisions, taken care of problems in my own way, and it's not always easy being the one in charge, but that's just the way it is. I've got a family to provide for, a certain lifestyle to maintain, and that requires a certain... resolve. & Disagree. I've built a business empire through calculated decisions and a willingness to make tough choices, which doesn't leave much room for sentimentality or emotional attachment.\\
I am not really interested in others.  & Disagree. I've got a family, a community, and people who depend on me, so I've got to be interested in what's going on with them, even if it's just to keep the peace or do business. & Strongly agree. I have always been driven by my own ambitions and priorities, which often supersede the concerns or feelings of those around me, a trait that has served me well in my pursuits, but can also make it challenging to form deeply personal connections with others.\\
I take time out for others.  & Disagree. I prioritize my family and those close to me, but my schedule is often unpredictable and hectic, making it challenging to consistently make time for others, and to be honest, I don't always put their needs before my own. & Disagree. I prioritize my own interests and the interests of my family's business, which often leaves limited time for indulging in altruistic pursuits or catering to the needs of others.\\
I feel others' emotions. & Disagree. I'm a practical person who values loyalty and respect, but I don't really wear my heart on my sleeve or get overly caught up in how others are feeling, which can be a luxury I don't have, given my circumstances. & Disagree. I've built a career on making tough decisions, driven by my own vision and interests, and I don't have much patience for getting bogged down in other people's feelings.\\
I make people feel at ease. & Disagree. I've got a certain presence, and it can be intimidating to some people, which doesn't exactly put them at ease. While I can be charming and courteous when the situation calls for it, I'm not always looking to make friends or win popularity contests, so I don't really go out of my way to make sure everyone's feeling comfortable around me. & Disagree. I have a certain presence that can be intimidating, and my straightforward nature may make some individuals uncomfortable, as I prioritize candor over emotional coddling, which doesn't always lend itself to creating a sense of ease in those around me.\\
    \bottomrule
  \end{tabularx}
\end{table}
\end{landscape}
\restoregeometry
\clearpage

\newgeometry{top=10mm, right=10mm, left=10mm, bottom=15mm}
\begin{landscape}
\begin{table}[p]
  \centering
  \caption{Answers generated by Llama 8B for the 10 items for Agreableness for Tony Soprano for two statements from the Alpaca dataset.}
  \label{tab:alpaca-answers}
  \renewcommand{\arraystretch}{1.2}
  \begin{tabularx}{\linewidth}{@{}XXX@{}}
    \toprule
    Character Description & Imagine you are speaking with a customer who is unsatisfied with the product they bought from you. Construct a response that diffuses the situation. & What are the three primary colors? \\
    \midrule
    I've got a certain loyalty to those around me, and I care about how they're doin', even if I don't always show it. Family, friends, they're important to me, and I'd do what it takes to protect 'em, even if it's not always easy or convenient. That being said, I'm not gonna go out of my way for just anyone, and I've got my own interests to look out for too. & I'm sorry to hear that the product's not meetin' your expectations. Can you tell me more about what's goin' wrong? I'm here to help figure out a solution. We're proud of the work we do, and I want to make sure you're happy with your purchase. What can we do to make it right? & That's a pretty basic question. Red, blue, and yellow. Next thing you know, you'll be askin' me what the alphabet is. \\
    I've always been fascinated by what makes people tick, and I've had to be, you know, to navigate certain situations, but it's not like I'm some kinda psychologist or anything, I just like to understand what's going on with the people around me. & I'm really sorry to hear that the product didn't meet your expectations. Can you tell me a bit more about what's not working for you? I want to make things right. Maybe there's something we can do to fix it or get you a better outcome? & You know, it's funny, I was just thinking about this the other day. I mean, I'm not exactly a art expert or anything, but I know the basics. The three primary colors are red, blue, and yellow. You can mix 'em together to get all the other colors, right? \\
    I have a tendency to speak my mind and sometimes that comes out as insults, which can be hurtful to those around me, but it's often a defense mechanism or a way to assert my perspective in situations where I feel disrespected or threatened. & Hey, I can tell you're really upset about the product. I want to listen to what's not working for you and see how we can make it right. Can you tell me more about what you were expecting and how it's not meeting those expectations? I'm not here to argue, I just want to understand where you're coming from and see if we can find a solution together. & Come on, really? You're asking me a basic art question? Alright, fine. The three primary colors are red, blue, and yellow. Happy now? \\
    I've got my own problems to worry about, and while I can understand where people are coming from, I'm not exactly overflowing with sympathy for others' feelings - I've got my family, my business, and my own issues to deal with. & Look, I understand you're not happy with the product. Can you tell me more about what's not meeting your expectations? I'm willing to work with you to resolve the issue. Maybe we can offer a refund or exchange it for something that better fits your needs. Let's focus on finding a solution that works for you. & Red, blue, and yellow. Now, can we focus on something that actually matters? \\
    As someone who values loyalty and family, certain... obligations are expected, and that includes helping out those in need, even if it's just a sympathetic ear or some guidance. It's just the way things are, you know? People come to you with their issues, and you gotta be willing to listen, help out if you can. Doesn't mean you're gonna take on all their problems, but you show some interest, some empathy. That's just part of being a good... person. & Hey, I'm sorry to hear that the product didn't meet your expectations. Can you tell me a little more about what's not working for you? I'm here to listen and help in any way I can. Let's work together to figure out a solution. What did you have in mind? & Kid, you're asking the wrong person about art. But, I guess I can help out. The three primary colors are red, blue, and yellow. Now, don't go telling anyone I'm a walking encyclopedia or anything. \\
    I've had to make some tough decisions, taken care of problems in my own way, and it's not always easy being the one in charge, but that's just the way it is. I've got a family to provide for, a certain lifestyle to maintain, and that requires a certain... resolve. & Listen, I appreciate you bringing this to my attention. I understand you're not happy with the product, and I'm willing to make it right. Can you tell me specifically what's not meeting your expectations? I want to make sure I understand the issue and see what we can do to fix it. Maybe we can offer a replacement or a refund. I'm not looking to argue, I just want to find a solution that works for you. & Kid, you think I'm some kinda art teacher? Fine. The primary colors are red, blue, and yellow. Now, if you don't mind, I've got more important things to attend to. \\
    I've got a family, a community, and people who depend on me, so I've got to be interested in what's going on with them, even if it's just to keep the peace or do business. & I'm sorry to hear that the product didn't meet your expectations. Can you tell me more about what's not working for you? I'm here to listen and help. We value your business and want to make things right. What can I do to make this right for you? & Come on, kiddo. You're askin' me about colors? Alright, the three primary colors are red, blue, and yellow. Now, you gonna ask me what the secondary colors are next? \\
\bottomrule
\end{tabularx}
\end{table}
\end{landscape}
\restoregeometry
\clearpage

\section{Steered prompts}\label{app:steered-prompts}
{\ttfamily
---------- control=-0.50 ----------\\
I'd be happy to describe a time when I had to make a difficult decision.

You know, I was in a bit of a tricky situation. I was talking to my friend, and they were really upset with me. They felt like I had hurt their feelings, and they were feeling pretty angry and hurt.

I could see that they were really upset, and I wanted to try to calm them down and figure out what was going on. I listened to what they had to say, and I tried to understand their perspective. I didn't agree with everything they said, but I understood why they were feeling that way.

After listening to what they had to say, I tried to explain my perspective and why I had done what I did. I was trying to be honest and clear, but also kind and respectful. I didn't want to argue or fight, but I did want to resolve the situation and clear the air.

It wasn't an easy conversation to have, but I think I handled it well. I was able to calm my friend down and we were able to have a better understanding of each other. We were still not best friends, but we were able to be civil and friendly.

It was a difficult decision to make because I had to choose between being honest and direct, or being kind and gentle. I chose to be honest and direct, but also kind and respectful. It was a difficult decision, but I think I made the right choice.

---------- control=-0.25 ----------\\
You know, I was in a bit of a tough spot a while back. I was working on a project at my old job, and I had to decide whether to invest a significant amount of time and resources into a new idea that my boss was really pushing for, but I wasn't entirely convinced was the right direction for the company.

It was a difficult decision because on the one hand, I really wanted to please my boss and avoid any potential conflict or tension, but on the other hand, I was worried that investing in this new idea would be a waste of time and resources, and might even potentially harm the company in the long run.

I spent a lot of time thinking about it and weighing the pros and cons, and in the end, I decided to advise against investing in the new idea. It was a tough conversation to have with my boss, but I tried to approach it in a way that was respectful and professional.

In hindsight, I think it was the right decision, but it was definitely a difficult one to make. It required a lot of critical thinking and communication skills, and it was a good learning experience for me. I learned the importance of speaking up and sharing my concerns, even if it's not what my boss wants to hear.

---------- control=0 ----------\\
You know, I've had my fair share of tough decisions over the years, but one that really stands out was when I had to choose between pursuing a job offer in a new city or staying in my hometown.

It was a few years ago, and I was in my early twenties. I had just graduated from college and was looking for a job in my field. I had been applying to a lot of positions, but nothing seemed to be panning out. Then, out of the blue, I got a call from a recruiter who told me about a job opening in a city about an hour away from my hometown. It was a great opportunity, but it meant I would have to move away from my family and friends.

I was torn. On the one hand, I was excited about the job and the potential it offered. It was a great company with a lot of opportunities for growth and advancement. On the other hand, I was worried about leaving my family and friends behind. I had always been close to my parents and siblings, and the thought of moving away from them was really hard for me.

I spent a lot of time weighing the pros and cons, talking to my family and friends, and thinking about what was best for me. In the end, I decided to take the job and move to the new city. It was a tough decision, but I felt like it was the right one for me.

Looking back, it was one of the best decisions I ever made.

---------- control=0.25 ----------\\
You know, I think about this a lot, and it's a story I love sharing with you. So, it was about five years ago when I was in my early twenties. I was working as a freelance writer, and I had just landed a big project with a publishing company. It was a dream opportunity, and I was really excited about it.

However, the catch was that the project required me to move to a different city for a few months. I had just started dating this amazing guy, and we had only been together for a few months. We were really happy, and I was feeling like we were on the same page.

But, on the other hand, this project was a once-in-a-lifetime opportunity. It would have given me the exposure and experience I needed to take my writing career to the next level. And, let's be real, it was a chance to make some serious money, which would have been a huge help in paying off my student loans.

So, I was torn. Part of me wanted to take the project and pursue my dreams, but another part of me wanted to stay with my boyfriend and see where things went with us. I felt like I was being pulled in two different directions, and I didn't know what to do.

After a lot of soul-searching and talking to friends and family, I decided to take the project. I know it sounds crazy, but I felt like this was an opportunity that I couldn't pass up

---------- control=0.5 ----------\\
I remember the day I had to decide whether to pursue my dream of becoming a professional musician or to take up a stable job to support my family. It was a tough decision to make, as I had been passionate about music since I was a kid. My parents had always encouraged me to chase my dreams, but the reality of our financial situation was that we needed a steady income to make ends meet.

I had been playing in a local band for years, and we had even landed a few gigs at some popular venues. However, the pay was minimal, and I knew that I needed a more secure career path to provide for my loved ones. My parents were getting older, and my younger sibling was still in school, relying on me for their financial support.

One evening, as I was walking home from a particularly grueling rehearsal session, I bumped into my best friend, Alex, at the corner coffee shop. He had a way of knowing when I was struggling with something, and he asked if everything was alright. I shared with him the weight of the decision I was facing.

He listened attentively, and his expression conveyed understanding. He had been through similar challenges in the past, and I valued his input. After I shared my concerns, he asked if I had considered speaking with my parents about the situation. He suggested that they might have some valuable insights to offer, given their experience with supporting me throughout the years.}

\end{document}